\documentclass[runningheads]{llncs}

\usepackage{eccv}

\usepackage{eccvabbrv}

\usepackage{graphicx}
\usepackage{booktabs}

\usepackage[dvipsnames]{xcolor}

\usepackage[accsupp]{axessibility}  

\usepackage{hyperref}

\usepackage{orcidlink}

\usepackage{multirow}

\usepackage{pifont}

\usepackage{colortbl} 

\usepackage{microtype}

\usepackage{algorithm}
\usepackage{algpseudocode}

\begin{document}

\title{Scal3R: Learning Efficient Multi-Relative Pose Query for Scalable Online 3D Reconstruction} 

\titlerunning{Scal3R}

\author{Chin-Yang Lin$^{1,2}$ \quad Yang-Che Sun$^{1}$ \quad Cheng Sun$^{2}$ \quad Fu-En Yang$^{2}$ \\ Min-Hung Chen$^{2}$ \quad Yen-Yu Lin$^{1}$ \quad Wei-Chen Chiu$^{1}$ \quad Yu-Lun Liu$^{1}$}

\authorrunning{C.-Y.~Lin et al.}

\institute{$^{1}$Department of Computer Science, National Yang Ming Chiao Tung University \quad $^{2}$NVIDIA}

\maketitle

\begin{figure*}[th]
    \centering
    \includegraphics[width=\textwidth]{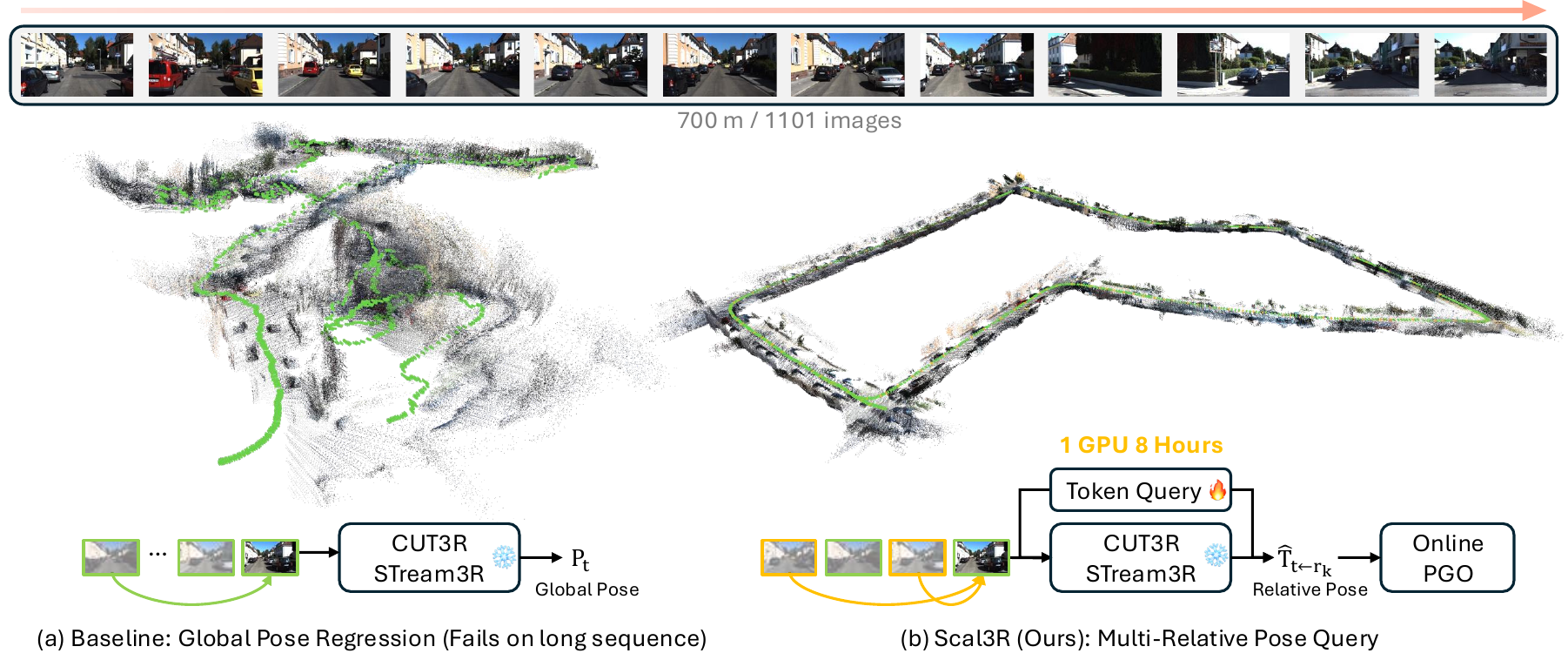} 
    \caption{
    \textbf{Scal3R enables scalable online 3D reconstruction on long video streams.} \textbf{(a)} Existing feed-forward models, such as CUT3R~\cite{wang2025continuous} and STream3R~\cite{lan2025stream3r}, regress absolute global poses ($P_t$) relative to a fixed first frame. This forces extrapolation far beyond its training distribution, resulting in catastrophic drift and geometric collapse on kilometer-scale sequences. \textbf{(b)} Scal3R reformulates the problem into a local multi-reference relative pose query ($\hat{\mathbf{T}}_{t \leftarrow r_k}$). By injecting lightweight learnable tokens into a frozen backbone and aggregating relative constraints via online Pose-Graph Optimization (PGO), Scal3R suppresses long-range drift and recovers globally consistent geometry. It requires only 8 hours of fine-tuning on a single GPU.
    }
\label{fig:teaser}
\end{figure*}

\begin{abstract}
  Online 3D reconstruction models perform poorly on long videos. 
  This happens because regressing poses relative to a fixed first-frame anchor forces extrapolation far beyond the training distribution.
  Small drifts accumulate and amplify into significant geometric collapse.
  However, we observe that per-frame depth remains stable throughout this failure.
  The backbone's local geometry remains intact; only the global pose head breaks down. 
  Motivated by this decoupling, we introduce \textbf{Scal3R}.
  This approach reformulates online reconstruction as multi-reference relative pose querying.
  We use lightweight learnable tokens, which make up about $\sim$1\% of the parameters, and inject them into a completely frozen backbone via asymmetric attention.
  This setup queries poses relative to multiple past keyframes.
  An online pose-graph optimization system with loop closure suppresses long-range drift. 
  Scal3R reaches convergence in 8 hours on a single GPU.
  It reduces the average ATE by over 60\% on KITTI compared to the online baseline.
  It also achieves state-of-the-art performance across Virtual KITTI, Sintel, TUM-Dynamic, ScanNet, and 7-Scenes.
  Project page: \url{https://linjohnss.github.io/scal3r/}
\end{abstract}

\section{Introduction}
\label{sec:intro}

Feed-forward models such as CUT3R~\cite{wang2025continuous} and STream3R~\cite{lan2025stream3r} enable real-time 3D reconstruction by decoding per-frame geometry into a unified global coordinate system via direct pose regression relative to the first frame. While this \textbf{global-anchor} paradigm works well for short sequences, it faces severe stability and scalability bottlenecks in long-range environments.

\begin{figure*}[t]
    \centering
    \includegraphics[width=\textwidth]{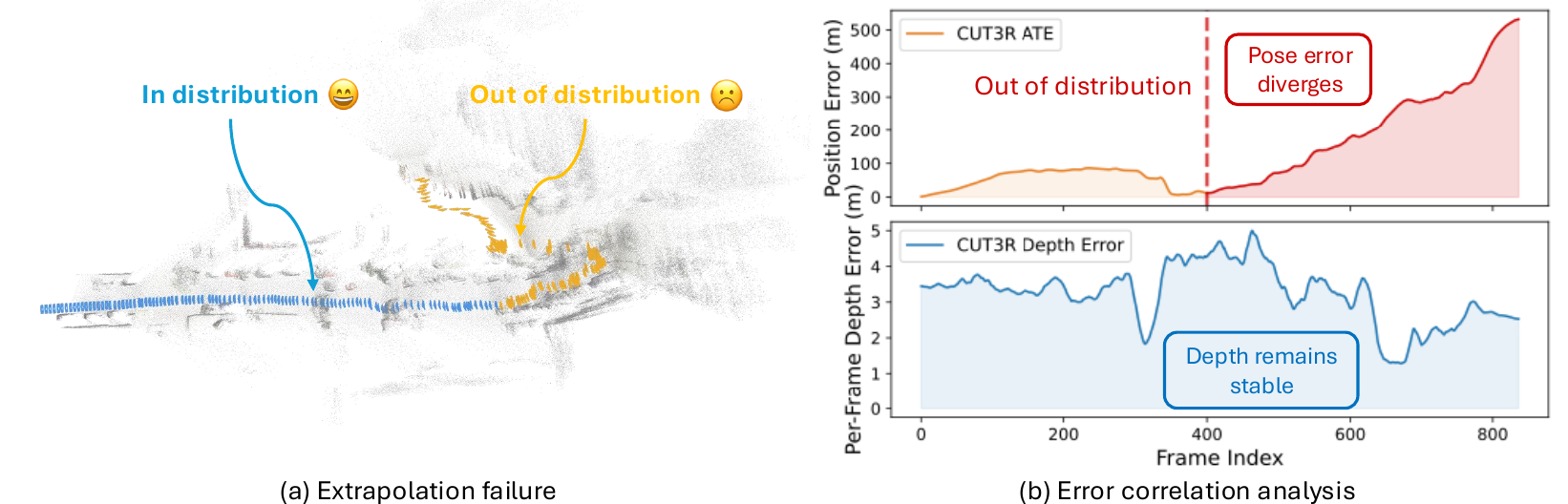} 
    \caption{
    \textbf{Global pose regression fails under out-of-distribution sequences, while local geometry remains reliable.}
    \textbf{(a)} Feed-forward models like CUT3R~\cite{wang2025continuous} produce accurate reconstructions for in-distribution frames (\textcolor{cyan}{\textbf{blue}}). However, they suffer from severe geometric collapse when extrapolating to unseen long-range trajectories (\textcolor{Dandelion}{\textbf{orange}}).
    \textbf{(b)} Our error correlation analysis reveals a critical decoupling. The global position error diverges catastrophically in the out-of-distribution region (\textcolor{red}{\textbf{red}}), while per-frame depth remains stable (\textcolor{NavyBlue}{\textbf{blue}}). This finding suggests that the backbone's local geometric representations are intact. This motivates Scal3R to freeze the base model and replace fragile global regression with stable multi-reference relative pose querying.
    }
    \label{fig:motivation}
\end{figure*}

This failure stems from two fundamental issues. 
First, existing 3D datasets~\cite{yeshwanth2023scannet++,wang2020tartanair} cover limited scene scales. This means that models trained on short sequences must extrapolate to global coordinates far outside their training distribution.
When deployed on real-world trajectories spanning hundreds of meters, even minor feature drift is amplified into geometric collapse (\cref{fig:motivation}a). Second, real-world camera motions are highly variable, and for unbounded video streams, the global coordinate system will inevitably encounter out-of-distribution trajectories, making feed-forward global pose regression theoretically unable to scale.

As shown in \cref{fig:motivation}b, this failure is highly localized: while global pose errors diverge catastrophically, per-frame depth remains consistently stable.
This indicates that the backbone's local geometric representations are intact and the failure is isolated to the global pose regression head. Motivated by this, we shift from \emph{global pose regression} to \emph{local relative querying}: rather than forcing the model to extrapolate a global mapping, we exploit its well-learned local geometry to estimate stable relative transformations via a visual query mechanism conditioned on reference viewpoints, with the backbone entirely frozen.

We present \textbf{Scal3R}, an efficient framework for scalable online 3D reconstruction (\cref{fig:teaser}). A small set of learnable tokens is injected into the frozen backbone via \emph{asymmetric attention}: pose tokens attend to image features as queries while image tokens compute self-attention exclusively among themselves, preserving the frozen representation space. Each pose token predicts the relative transformation between the current frame and a historical reference, constraining localization to the local viewpoint domain where the model is most reliable.

To maintain global consistency, Scal3R incorporates multi-reference relative querying and online pose-graph optimization: relative poses are queried against multiple dynamically selected reference frames simultaneously, aggregated via PGO into a drift-free trajectory, and supplemented by a loop closure mechanism that integrates naturally into the multi-reference pipeline. As shown in \cref{fig:teaser}, Scal3R produces geometrically consistent reconstructions on sequences spanning hundreds of meters, finetuned in only 8 hours on a single GPU.

In summary, our contributions are as follows:
\begin{itemize}
\item We identify that global extrapolation instability and limited training data coverage cause long-sequence collapse in long-sequence reconstruction. We also show that local geometric representations remain reliable throughout.
\item We propose Scal3R. It introduces multi-reference relative pose querying via visual prompt tuning on a frozen backbone. Asymmetric attention injection ensures that pose learning does not degrade point cloud quality.

\item Integrating multi-reference querying with online PGO, Scal3R achieves low drift on kilometer-scale sequences. It converges in about 8 hours on a single GPU using only 4-view training samples.
\end{itemize}

\vspace{-.5em}
\section{Related Work}

\vspace{-.5em}
\subsubsection{Multi-view 3D Reconstruction.}
Early reconstruction relied on offline Structure-from-Motion~\cite{schoenberger2016sfm,pan2024global} and Multi-View Stereo~\cite{schoenberger2016mvs} pipelines.
Optimization-based approaches jointly refine camera poses with the scene representation~\cite{liu2023robust,chen2024improving,meuleman2023progressively,lin2025longsplat}, but remain per-scene and offline.
Feed-forward methods dramatically improved efficiency by predicting geometry in a single forward pass~\cite{Wang_2024_CVPR,duisterhof2025mastrsfm,mast3r_arxiv24,zhang2024monst3r,chen2025easi3r,jang2025pow3r,du2025moonseg3r,sun20263am,su2024boostmvsnerfs,charatan2024pixelsplat,chen2024mvsplat,ye2024no},
with recent transformer-based models scaling to large unordered collections via joint pose-and-geometry prediction~\cite{wang2025vggt,wang2024vggsfm,wang2025pi} and efficient aggregation strategies~\cite{wang2025faster,shen2025fastvggt,wang2025flashvggt,elflein2025light3r,chen2025co,sun2025avggt,elflein2026vgg,wang2025amb3r,zhang2025d4rt,yang2025fast3r,tang2025mv,lin2025depth,keetha2025mapanything,cong2026flow3r,liu2025regist3r,sun2025uni3r,lin2024frugalnerf}.
A complementary trend adapts frozen 3D foundation models for downstream tasks without backbone retraining~\cite{smart2024splatt3r,jena2025sparfels,hsiao2026vecset}.
Scal3R follows this paradigm but uniquely targets scalable pose estimation on unbounded video streams, where batch-processing systems fundamentally cannot operate.

\vspace{-.5em}
\subsubsection{Online 3D Reconstruction.}
Online methods shift from batch processing to incremental inference, processing video frame by frame, achieved through recurrent TSDF fusion~\cite{sun2021neuralrecon}, differentiable bundle adjustment~\cite{teed2021droid}, and progressive radiance-field optimization~\cite{meuleman2023progressively}.
CUT3R~\cite{wang2025continuous} and STream3R~\cite{lan2025stream3r} bring this to 3D foundation models via persistent state updates and causal Transformers, spawning a broad family of streaming systems~\cite{shen2025grs,wang20243d,zheng2026ttsa3r,khafizov2025g,streamVGGT,cabon2025must3r,wu2025point3r,mahdi2025evict3r,antsfeld2026s,yuan2026infinitevggt,taher2025kv,li2025wint3r,chen2025long3r} and SLAM integrations~\cite{murai2025mast3r,maggio2026vggt,maggio2025vggt,slam-former,liu2025slam3r,zhang2025vista,ding2025laser,huang2025vipe,yugay2025visualodometrytransformers}.
However, all share a critical flaw in that poses are regressed relative to the first frame, anchoring the trajectory to a single global reference.
As shown in \cref{fig:teaser,fig:motivation}, this strategy becomes increasingly fragile as sequences grow, where small feature drifts are amplified into catastrophic geometric collapse.

\vspace{-.5em}
\subsubsection{Long-sequence Streaming 3D Reconstruction.}
Suppressing drift over kilometer-scale sequences remains an open challenge, addressed through test-time gradient updates~\cite{chen2025ttt3r}, training-free memory management~\cite{yuan2026infinitevggt,zheng2026ttsa3r}, long-range token pools~\cite{li2025wint3r}, explicit spatial memory~\cite{wu2025point3r}, stage-decoupled streaming~\cite{cheng2026longstreamlongsequencestreamingautoregressive}, and offline global optimization~\cite{deng2025vggtlongchunkitloop,xiong2026vggt,dai2026keyframe}, each trading off online capability against global consistency.
Earlier per-scene methods handle long casual videos by incrementally estimating poses with learned 3D priors~\cite{lin2025longsplat} or by progressively allocating local radiance fields rather than a single global representation~\cite{meuleman2023progressively}, an early departure from single-anchor formulations.
A unifying insight from the visual odometry literature is that relative formulations generalize better than absolute ones~\cite{chen2024leap,dong2025reloc3r}.
This insight guides Scal3R. Rather than improving global-anchor regression, we reformulate the problem as multi-reference relative pose querying on a frozen backbone, eliminating the root extrapolation failure with only ${\sim}$1\% additional parameters.

\vspace{-.5em}
\subsubsection{Efficient Prompt Tuning.}
Parameter-efficient adaptation has shown that frozen pre-trained models need very little change to transfer well. Adapters~\cite{houlsby2019parameter}, prefix tokens~\cite{li2021prefix}, soft prompts~\cite{lester2021power}, low-rank perturbations~\cite{hu2022lora}, and parallel adapter modules~\cite{chen2022adaptformer} all match or exceed full fine-tuning at under 2\% of parameters, a finding confirmed broadly across vision transformers~\cite{xin2024parameter,jia2022visual,tu2023visual,huang2025cvpt,ren2025vpt}.
This paradigm has since reached 3D vision, where geometry-aware prompts and low-rank adapters on frozen 3D transformers~\cite{tang2024point,ai2025gaprompt,zhou2024dynamic,wang2025pointlora} and reconstruction backbones~\cite{lu2024lora3d,wang2026moe3d} consistently match full fine-tuning, and attention-level token gating on a frozen large reconstruction model enables mesh editing without backbone retraining~\cite{hsiao2026vecset}.
In 3D reconstruction, Human3R~\cite{chen2025human3r} first demonstrates prompt tuning on a frozen CUT3R for joint human-scene reconstruction.
Scal3R is the first to apply this paradigm to relative pose estimation, recasting it as a multi-reference prompt query task via asymmetric attention injection that preserves the backbone's pointmap quality while gaining globally consistent motion representations.

\section{Method}
\label{sec:Method}

\begin{figure*}[t]
    \centering
    \includegraphics[width=\textwidth]{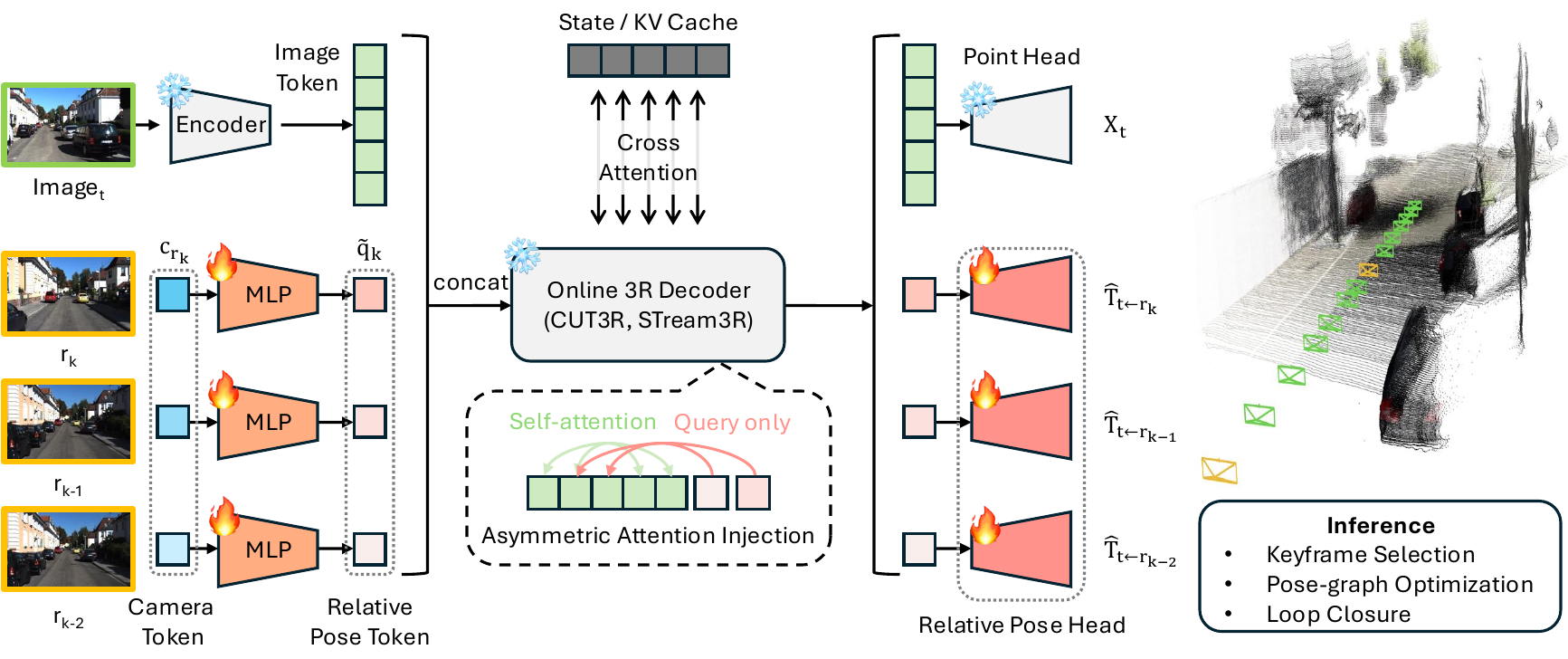} 
    \caption{
    \textbf{Overview of the Scal3R framework.}
    For an incoming frame $\text{Image}_t$, a frozen encoder extracts dense image tokens. At the same time, historical camera tokens from selected reference frames ($r_k, r_{k-1}, \dots$) are projected using lightweight trainable MLPs to generate relative pose tokens. These tokens are concatenated and sent into a completely frozen 3D reconstruction decoder (\emph{e.g.}, CUT3R~\cite{wang2025continuous} or STream3R~\cite{lan2025stream3r}). Our \textbf{Asymmetric Attention Injection} mechanism is crucial as it ensures that relative pose tokens serve only as queries to extract geometric cues, while image tokens perform self-attention exclusively among themselves. This method preserves the original high-quality point cloud generation ($X_t$) through the frozen Point Head, while the trainable Relative Pose Head predicts robust multi-reference relative transformations ($\hat{\mathbf{T}}_{t \leftarrow r_k}$). Finally, an online inference backend (PGO and loop closure) aggregates these local constraints to produce a globally consistent trajectory.
    }
    \label{fig:pipeline}
\end{figure*}

\subsection{Overview}
\label{sec:overview}

Scal3R addresses online 3D reconstruction by reformulating global pose regression as a multi-reference relative pose query problem. Given a streaming sequence of images $\mathcal{I} = \{I_1, I_2, \dots, I_T\}$, instead of directly regressing absolute camera poses in a unified world coordinate system, we query relative poses with respect to a set of maintained reference frames. This reformulation fundamentally eliminates the long-horizon extrapolation instability that plagues existing global-regression approaches.

Architecturally, Scal3R builds upon frozen pretrained online 3D reconstruction backbones (\eg, CUT3R~\cite{wang2025continuous} or STream3R~\cite{lan2025stream3r}), preserving their rich spatiotemporal geometric priors. A lightweight set of learnable tokens is injected into the frozen decoder via an asymmetric attention mechanism, enabling relative pose decoding without modifying the pretrained weights. At the backend, an online Pose-graph Optimization (PGO) module aggregates the predicted pairwise relative constraints into a globally consistent trajectory. An overview of the full pipeline is shown in \cref{fig:pipeline}.

\subsection{Preliminaries: Online 3D Reconstruction Backbones}
\label{sec:preliminaries}
We briefly review the two representative backbone paradigms underlying Scal3R.

\subsubsection{Persistent state model (CUT3R).} At each timestep $t$, the frozen online 3D reconstruction backbone processes the current frame $I_t$ together with a persistent hidden state $S_{t-1}$ encoding the scene history, producing an updated state $S_t$ and local geometry prediction $G_t$.

\subsubsection{Causal Transformer model (STream3R).} At each timestep $t$, the frozen online 3D reconstruction backbone processes the current frame $I_t$ via causal attention over a sliding feature window $\{F_{t-k}, \dots, F_t\}$ to perform cross-temporal geometric alignment in feature space, producing local geometry prediction $G_t$.

Both paradigms share a common decoding structure. At each frame, the decoder maintains image feature tokens $F_t \in \mathbb{R}^{H \times W \times C}$ and a dedicated camera token $\mathbf{c}_t \in \mathbb{R}^D$. Pointmaps are decoded from $F_t$ for local geometry, while the global camera pose $P_t \in SE(3)$ relative to the first frame is regressed from $\mathbf{c}_t$.

Although effective for short sequences, global-reference regression degrades over long sequences: as the sequence grows, the model must align each new frame to an increasingly distant first-frame coordinate system, causing small feature drifts to be amplified into severe geometric collapse at the decoding stage. Scal3R retains the rich representations $F_t$ and $\mathbf{c}_t$ learned by these backbones, while discarding their unstable global pose regression heads. Instead, we leverage historical camera tokens $\{\mathbf{c}_{r_k}\}$ stored in the pose token buffer as geometric conditioning signals to enable scalable relative pose queries (\cref{sec:prompt_tuning}).




\subsection{Multi-Reference Relative Pose Tuning}
\label{sec:prompt_tuning}
Our core contribution is a parameter-efficient prompt tuning mechanism that enables robust multi-reference relative pose prediction on a completely frozen backbone. The total number of newly introduced parameters accounts for $\sim$1\% of the backbone's total parameter count.
To endow the model with the ability to query multiple reference viewpoints simultaneously, we maintain a \emph{pose token buffer} for storing the camera tokens of selected past keyframes. We learn a shared base query token $\mathbf{q} \in \mathbb{R}^D$ that serves as a query template directing the decoder to extract the geometric relationship between the current frame and a given reference frame. For each reference slot $k$, the corresponding reference frame features retrieved from the buffer are projected into feature space via a lightweight MLP and fused with the base token by additive injection:
\begin{equation} \small
    \tilde{\mathbf{q}}_k = \mathbf{q} + \mathrm{MLP}(\mathbf{c}_{r_k}),
\end{equation}
where $\mathbf{c}_{r_k}$ is the camera token of the $k$-th reference frame. This dynamic assembly allows the system to flexibly scale the number of active queries $K$ based on available references, ensuring robustness during sequence initialization or buffer resets. Importantly, since each token queries independently, the number of reference frames can be freely extended at inference time without retraining.

\subsubsection{Asymmetric Attention Injection.}
\label{sec:asymmetric_attn}
\begin{figure*}[t]
    \centering
    \begin{minipage}[t]{0.48\textwidth}
        \centering
        \includegraphics[width=\textwidth]{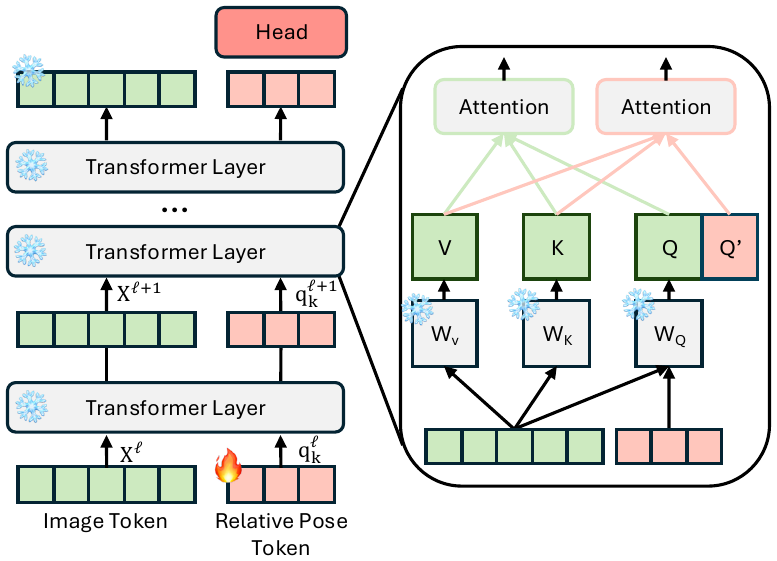}
        \caption{\textbf{Asymmetric Attention Injection.} Relative pose tokens are injected only 
        as additional queries ($Q'$), without modifying the keys and values of image tokens. This asymmetric design enables pose conditioning while preserving the pretrained image 
        representations intact.}
        \label{fig:token_query}
    \end{minipage}
    \hfill
    \begin{minipage}[t]{0.48\textwidth}
        \centering
        \includegraphics[width=\textwidth]{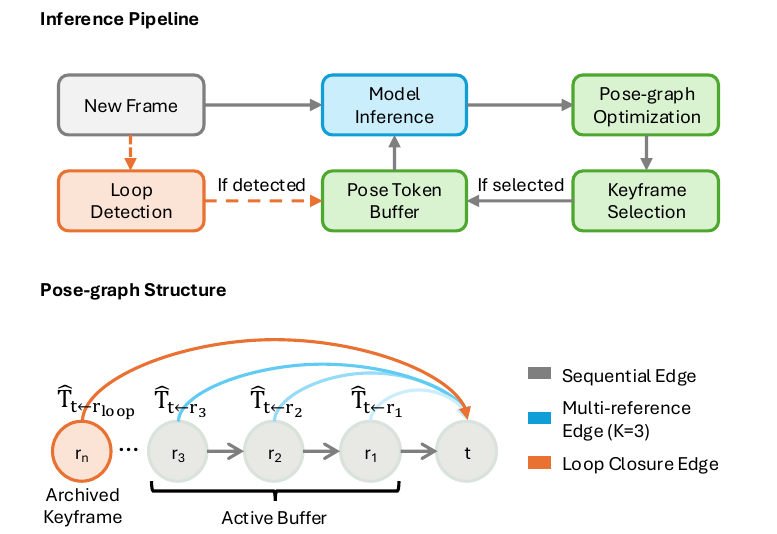}
        \caption{\textbf{Inference pipeline and pose graph structure.} To suppress accumulative drift in long sequences, incoming frames are registered via keyframe selection and pose-graph optimization over sequential, multi-reference ($K{=}3$), and loop closure edges before being committed to the pose token buffer.}
        \label{fig:online_pgo}
    \end{minipage}
\end{figure*}

Naively inserting new tokens into the decoder's self-attention would perturb the attention distribution of image tokens, degrading pointmap reconstruction quality. We instead propose \emph{asymmetric attention injection} (\cref{fig:token_query}), where the pose query tokens $\{\tilde{\mathbf{q}}_k\}$ participate in decoder attention exclusively as \emph{queries}, attending to all image tokens to extract geometric features, while image tokens compute their Keys and Values without attending to the pose query tokens. For a decoder layer with image tokens $\mathbf{X}$:
{
\small
\begin{align}
    \tilde{\mathbf{q}}_k^{\ell+1} &= \mathrm{Attention}\!\left(\mathbf{Q}=\tilde{\mathbf{q}}_k^\ell,\ \mathbf{K}=\mathbf{X}^\ell,\ \mathbf{V}=\mathbf{X}^\ell\right), \\
    \mathbf{X}^{\ell+1} &= \mathrm{SelfAttention}\!\left(\mathbf{X}^\ell\right).
\end{align}
}
This one-directional information flow guarantees that the image feature representation space remains identical to that of the original frozen model, fully preserving pointmap reconstruction fidelity. No attention mask is needed, as pose tokens never enter the image K/V sequence.

\subsubsection{Relative Pose Decoding and Loss.}
\label{sec:pose_decoding}
After multi-layer feature exchange, each pose query token $\tilde{\mathbf{q}}_k$ encapsulates the relative geometric constraint between the current frame $t$ and its corresponding reference frame $r_k$. A lightweight MLP head decodes these tokens into relative poses. We adopt the 6D rotation representation~\cite{zhou2019continuity} to ensure continuity in the rotation space, and output the relative transformation:
\begin{equation} \small
    \hat{\mathbf{T}}_{t \leftarrow r_k} = \mathrm{Head}\!\left(\tilde{\mathbf{q}}_k\right) \in SE(3).
\end{equation}
The training loss supervises rotation $\mathbf{R}$ and translation $\mathbf{t}$ separately, where $\mathbf{R}_{pred}^k$ and $\mathbf{t}_{pred}^k$ denote the rotation matrix and translation vector decomposed from $\hat{\mathbf{T}}_{t \leftarrow r_k}$, and $\mathbf{R}_{gt}^k$, $\mathbf{t}_{gt}^k$ are the corresponding ground-truth components. To handle monocular scale ambiguity, translation vectors are scale-aligned before loss computation. The total loss aggregates over all $K$ reference links:
\begin{equation} \small
    \mathcal{L} = \sum_{k=1}^{K} \left( \lambda_R \left\|\mathbf{R}_{pred}^k - \mathbf{R}_{gt}^k\right\|_F + \lambda_t \left\|\mathbf{t}_{pred}^k - \mathbf{t}_{gt}^k\right\|_2 \right),
\end{equation}
where $\lambda_R$ and $\lambda_t$ are loss weighting hyperparameters.

\subsection{Online Pose-graph Optimization}
\label{sec:pgo}

While multi-reference relative pose predictions provide accurate pairwise constraints, naively 
chaining them accumulates drift over long sequences. We therefore integrate an online 
pose-graph optimization (PGO) framework (\cref{fig:online_pgo}) that uses the predicted 
relative poses as between-factors and performs incremental trajectory correction as new frames arrive.

\subsubsection{Keyframe Selection.}
\label{sec:keyframe}
Including every frame in the pose graph introduces numerical redundancy and unnecessary computation. We adopt an online 3D overlap-based keyframe selection strategy. A KD-tree spatial index maintains the reconstructed 3D point cloud; the visible point set for each frame is determined by projecting predicted 3D points onto a unit sphere and computing the angular overlap with past keyframes, following~\cite{cabon2025must3r}, to avoid interference from geometrically non-adjacent regions. For each incoming frame $t$, the predicted 3D points are transformed to world coordinates via the current pose estimate. If the depth-normalized overlap score falls below a threshold $\tau_\text{overlap}$ and the median depth confidence exceeds $\tau_\text{conf}$, the frame is designated as a keyframe, indicating novel geometry with reliable prediction quality. Only keyframes update the frozen decoder's KV cache and enter the pose token buffer. Non-keyframe KV states are discarded by restoring the pre-forward snapshot, keeping the streaming decoder state clean. The first $N_\text{init}$ frames are unconditionally treated as keyframes to initialize the system.

\subsubsection{Pose-graph Optimization.}
\label{sec:pgo_detail}
We model the trajectory as a factor graph where each camera pose $T_t \in SE(3)$ is a variable node. The multi-reference relative poses form between-factors connecting the current frame to its $K_t$ references:
\begin{equation} \small
    E_\text{between} = \sum_{t} \sum_{k=1}^{K_t} \rho\!\left(\left\| \log\!\left( T_{r_k}^{-1} T_t \cdot \hat{T}_{t \leftarrow r_k} \right) \right\|^2_{\Sigma_{t,k}} \right),
\end{equation}
where $\hat{T}_{t \leftarrow r_k}$ is the predicted relative pose, $\Sigma_{t,k}$ is a diagonal noise covariance, and $\rho(\cdot)$ is the Huber robust kernel to downweight outlier constraints. To account for higher uncertainty in predictions between temporally distant frame pairs, we adopt a gap-dependent noise model where the standard deviation scales as $\sigma = \sigma_\text{base} \cdot \Delta^{0.5}$ with frame gap $\Delta$. We employ iSAM2~\cite{kaess2012isam2} for incremental optimization. Upon each new frame arrival, the factor graph is updated and efficiently re-optimized via the Bayes tree structure. Optimized poses are written back to the buffer so that subsequent frames use corrected references.

For long sequences, the frozen decoder's streaming state is reset every $N_\text{reset}$ frames to prevent memory overflow and feature degradation. To maintain pose-graph connectivity across resets, the last frame of each segment is re-fed as the first frame of the next segment, with a tight identity constraint imposed between the two corresponding nodes in the factor graph.

\subsubsection{Loop Closure.}
\label{sec:loop_closure}
Despite PGO continuously correcting local drift, long-term trajectory consistency requires explicitly detecting and closing loops when the camera revisits previously observed regions. A key advantage of our multi-reference design is that loop closure integrates naturally into the existing inference pipeline. When a loop candidate is detected between the current frame $t$ and a past keyframe $r_\text{loop}$, the archived camera token of $r_\text{loop}$ is simply re-injected into the pose token buffer as an additional reference slot. The frozen model then predicts a long-range relative pose constraint $\hat{\mathbf{T}}_{t \leftarrow r_\text{loop}}$ without any architectural modification, which is added to the pose graph as a high-confidence edge with a tight Gaussian noise model.

For loop detection, we employ a pretrained DINOv2~\cite{oquab2023dinov2} backbone with a SALAD aggregation layer~\cite{izquierdo2024optimal} to produce discriminative scene-level descriptors, indexed online via FAISS over keyframes only. Candidates are filtered by cosine similarity threshold $\tau_\text{sim}$, minimum temporal gap $\tau_\text{gap}$, and non-maximum suppression within a window $w_\text{nms}$ to suppress redundant detections.

\section{Experiments}
\label{sec:Experiments}

\subsection{Implementation Details}

\subsubsection{Model Configurations.}
We build upon two representative online 3D reconstruction backbones: CUT3R, which centers on persistent state updates, and STream3R, which is based on causal Transformers. In our experiments, we employ their 24-layer Transformer backbones (comprising a DINOv2 encoder and a Transformer decoder) and keep them entirely frozen to leverage their strong spatiotemporal geometric priors. For each incoming frame, we introduce a set of lightweight, learnable relative pose query tokens, which account for only approximately 1\% of the total model parameters. These tokens are injected into the decoder layers via asymmetric attention injection to extract geometric constraints of the current frame relative to the reference frames in the pose token buffer. A pose decoding head then maps these features into $SE(3)$ space, predicting the 6D rotation and translation vectors.

\subsubsection{Training.}
We fine-tune our model on the TartanAir~\cite{wang2020tartanair} dataset, which provides diverse scenarios and precise trajectory ground truth. To ensure robustness to varying motion velocities and baseline lengths, we adopt a Random Interval Sampling strategy during training. For each training sample, we select 4 views from a sequence, with one serving as the current frame $I_t$ and the remaining three as reference frames ($K=3$), and randomly perturb the temporal intervals between frames. This mechanism forces the model to extract stable relative pose representations under varying levels of geometric constraint. Since each pose query token attends independently, the number of reference frames can be freely scaled at inference time without retraining; we use $K=12$ during inference. For the long outdoor benchmarks (KITTI and vKITTI), we reset the frozen decoder's streaming state every $N_{\text{reset}}=10$ frames; all other datasets use no reset. We train with a batch size of 8 for 40 epochs using the AdamW optimizer with a learning rate of $1 \times 10^{-4}$. Thanks to the frozen backbone and lightweight query tokens, the entire fine-tuning converges in approximately 8 hours on a single NVIDIA A100 GPU, avoiding the collapse of geometric priors commonly observed in full-parameter fine-tuning on small-scale datasets.

\subsubsection{Baselines.}
We compare Scal3R with offline transformers, streaming models, and SLAM-style systems. Offline transformers include VGGT~\cite{wang2025vggt}, $\pi^3$~\cite{wang2025pi}, Fast3R~\cite{yang2025fast3r}, and DA3~\cite{lin2025depth}. Streaming baselines include CUT3R~\cite{wang2025continuous}, MUSt3R~\cite{cabon2025must3r}, TTT3R~\cite{chen2025ttt3r}, STream3R~\cite{lan2025stream3r}, WinT3R~\cite{li2025wint3r}, StreamVGGT~\cite{streamVGGT}, and Point3R~\cite{wu2025point3r}. MASt3R-SLAM~\cite{murai2025mast3r} is included as an incremental SLAM counterpart. All methods operate in an intrinsic-free setting, taking only RGB input without known camera intrinsics, and are evaluated with official default settings under a unified protocol.

\begin{table}[t]
  \caption{\textbf{Camera pose estimation on KITTI (ATE $\downarrow$).} The upper block lists offline baselines, and the lower block reports online methods. Scal3R reduces average ATE by over $60\%$ compared to the strongest online competitor TTT3R, with particularly pronounced gains on long-range sequences. \textbf{-} denotes OOM or tracking failure.}
  \label{tab:kitti_final_refined}
  \centering
\setlength{\tabcolsep}{3pt}
  \resizebox{\textwidth}{!}{
  \begin{tabular}{l ccccccccccc | c}
    \toprule
    \multirow{3}{*}{\textbf{Methods}} & \multicolumn{11}{c}{\textbf{KITTI (ATE $\downarrow$)}} & \multirow{3}{*}{\textbf{Avg.}} \\
    \cmidrule(lr){2-12}
    & 00 & 01 & 02 & 03 & 04 & 05 & 06 & 07 & 08 & 09 & 10 & \\
    & \scriptsize{4542$\times$} & \scriptsize{1101$\times$} & \scriptsize{4661$\times$} & \scriptsize{801$\times$} & \scriptsize{271$\times$} & \scriptsize{2761$\times$} & \scriptsize{1101$\times$} & \scriptsize{1101$\times$} & \scriptsize{4071$\times$} & \scriptsize{1591$\times$} & \scriptsize{1201$\times$} & \\
    & \scriptsize{3.7km} & \scriptsize{2.5km} & \scriptsize{5.1km} & \scriptsize{0.6km} & \scriptsize{0.4km} & \scriptsize{2.2km} & \scriptsize{1.2km} & \scriptsize{0.7km} & \scriptsize{3.2km} & \scriptsize{1.7km} & \scriptsize{0.9km} & \\
    \midrule
    VGGT \cite{wang2025vggt} & - & - & - & - & - & - & - & - & - & - & - & - \\
    Fast3R \cite{yang2025fast3r} & - & 723.1 & - & 166.9 & 112.2 & - & 137.4 & 90.9 & - & 225.6 & 211.7 & 238.3 \\
    DA3 \cite{lin2025depth} & - & - & - & - & 12.6 & - & - & - & - & - & - & 12.6 \\
    $\pi^3$ \cite{wang2025pi} & - & - & - & - & 2.3 & - & - & - & - & - & - & 2.3 \\
    \midrule
    MASt3R-SLAM \cite{murai2025mast3r} & 188.5 & 562.9 & 282.4 & 121.7 & 92.6 & - & 57.2 & 77.0 & 263.6 & 184.1 & 179.1 & 200.9 \\
    MUSt3R \cite{cabon2025must3r} & - & 490.1 & - & 121.5 & 58.1 & - & 100.6 & 66.8 & - & - & - & 167.4 \\
    CUT3R \cite{wang2025continuous} & 191.0 & 644.1 & 295.4 & 150.1 & 20.1 & 155.6 & 132.7 & 72.8 & 231.8 & 206.2 & 179.6 & 207.2 \\
    Point3R \cite{wu2025point3r} & - & - & - & - & - & - & - & - & - & - & - & - \\
    STream3R \cite{lan2025stream3r} & 189.0 & 694.9 & 302.6 & 162.7 & 100.5 & 159.1 & 122.7 & 87.1 & 264.5 & 220.9 & 195.9 & 227.3 \\
    WinT3R \cite{li2025wint3r} & - & 698.1 & - & 144.5 & 89.4 & 150.8 & 135.8 & 76.0 & 242.8 & 200.9 & 210.3 & 216.5 \\
    StreamVGGT \cite{streamVGGT} & - & - & - & - & 98.8 & - & - & - & - & - & - & 98.8 \\
    TTT3R \cite{chen2025ttt3r} & 178.4 & 529.1 & 280.8 & 98.0 & 11.4 & 147.9 & 132.1 & 70.2 & 240.2 & 191.0 & 125.2 & 182.2 \\
    \midrule
    \textbf{Ours (CUT3R)}    & \cellcolor{red!25}45.3  & \cellcolor{red!25}165.0 & \cellcolor{red!25}139.9 & \cellcolor{orange!25}33.9 & \cellcolor{orange!25}9.9  & \cellcolor{red!25}29.4 & \cellcolor{orange!25}47.4 & \cellcolor{red!25}6.4  & \cellcolor{orange!25}227.0 & \cellcolor{red!25}29.6 & \cellcolor{red!25}33.2 & \cellcolor{red!25}69.7 \\
    \textbf{Ours (STream3R)} & \cellcolor{orange!25}57.9 & \cellcolor{orange!25}176.4 & \cellcolor{orange!25}170.8 & \cellcolor{red!25}10.9 & \cellcolor{red!25}9.3 & \cellcolor{orange!25}39.1 & \cellcolor{red!25}18.1 & \cellcolor{orange!25}15.2 & \cellcolor{red!25}173.8 & \cellcolor{orange!25}73.9 & \cellcolor{orange!25}33.9 & \cellcolor{orange!25}70.8 \\
    \bottomrule
  \end{tabular}
  }
\end{table}

\begin{table*}[t]
  \centering
  \begin{minipage}[t]{0.58\textwidth}
    \centering
    \caption{\textbf{Camera pose estimation on Virtual KITTI (ATE $\downarrow$).} The upper block lists offline baselines, the middle block reports online methods, and the lower block presents our Scal3R variants. Scal3R surpasses all streaming methods by a large margin and approaches the accuracy of offline approaches on long-range sequences. \textbf{-} denotes OOM or tracking failure.}
    \label{tab:vkitti_final_refined}
\setlength{\tabcolsep}{3pt}
    \resizebox{\textwidth}{!}{
    \begin{tabular}{l ccccc | c}
      \toprule
      \multirow{2}{*}{\textbf{Methods}} & \multicolumn{5}{c}{\textbf{vKITTI (ATE $\downarrow$)}} & \multirow{2}{*}{\textbf{Avg.}} \\
      \cmidrule(lr){2-6}
      & \begin{tabular}[c]{@{}c@{}}Scene01\\ \scriptsize{447x, 1.3km}\end{tabular}
      & \begin{tabular}[c]{@{}c@{}}Scene02\\ \scriptsize{233x, 0.6km}\end{tabular}
      & \begin{tabular}[c]{@{}c@{}}Scene06\\ \scriptsize{270x, 0.6km}\end{tabular}
      & \begin{tabular}[c]{@{}c@{}}Scene18\\ \scriptsize{339x, 1.1km}\end{tabular}
      & \begin{tabular}[c]{@{}c@{}}Scene20\\ \scriptsize{837x, 2.5km}\end{tabular} & \\
      \midrule
      VGGT \cite{wang2025vggt}               & -      & 0.22  & -     & -      & -       & 0.22  \\
      Fast3R \cite{yang2025fast3r}            & 72.71 & 23.72 & 6.05 & 58.93 & 159.40 & 64.16 \\
      DA3 \cite{lin2025depth}                 & 25.22 & 0.22  & 0.17 & 12.53 & 182.89 & 44.20 \\
      $\pi^3$ \cite{wang2025pi}               & 4.63  & 0.33  & 0.18 & 3.04  & -       & 2.04  \\
      \midrule
      MASt3R-SLAM \cite{murai2025mast3r}      & 75.16 & -      & 7.056 & -      & -       & 41.11 \\
      MUSt3R \cite{cabon2025must3r}           & 77.23 & 7.39  & \cellcolor{red!25}0.28 & 48.53 & 170.91 & 60.87 \\
      CUT3R \cite{wang2025continuous}         & 59.76 & 29.73 & 0.65 & 44.89 & 146.92 & 56.39 \\
      Point3R \cite{wu2025point3r}            & 75.70 & 30.44 & 4.81 & 62.85 & 138.24 & 62.41 \\
      STream3R \cite{lan2025stream3r}         & 66.83 & 25.47 & 1.52 & 68.18 & 223.07 & 77.01 \\
      WinT3R   \cite{li2025wint3r}                               & 59.02 & 28.61 & 1.10 & 42.89 & 203.16 & 66.96 \\
      StreamVGGT \cite{streamVGGT}            & 55.09 & 32.91 & 0.57 & 57.18 & -       & 36.44 \\
      TTT3R \cite{chen2025ttt3r}              & 29.05 & 12.15 & \cellcolor{orange!25}0.54 & 6.75  & 77.90  & 25.28 \\
      \midrule
      \textbf{Ours (CUT3R)}    & \cellcolor{red!25}4.50 & \cellcolor{red!25}0.72 & 0.66 & \cellcolor{red!25}5.56 & \cellcolor{red!25}16.72 & \cellcolor{red!25}5.63 \\
      \textbf{Ours (STream3R)} & \cellcolor{orange!25}5.39 & \cellcolor{orange!25}3.26 & 3.21 & \cellcolor{orange!25}6.40 & \cellcolor{orange!25}21.32 & \cellcolor{orange!25}7.92 \\
      \bottomrule
    \end{tabular}
    }
  \end{minipage}
  \hfill
  \begin{minipage}[t]{0.4\textwidth}
    \centering
    \caption{\textbf{Camera pose estimation on Sintel, TUM-Dynamic, and ScanNet (ATE $\downarrow$).} Scal3R achieves state-of-the-art performance across all three benchmarks, demonstrating strong generalization to diverse unseen indoor and synthetic environments.}
    \label{tab:sintel_tum_scannet_results}
\setlength{\tabcolsep}{3pt}
    \resizebox{\textwidth}{!}{
    \begin{tabular}{l ccc}
      \toprule
      \multirow{2}{*}{\textbf{Methods}} & \textbf{Sintel} & \textbf{TUM} & \textbf{ScanNet} \\
      \cmidrule(lr){2-2} \cmidrule(lr){3-3} \cmidrule(lr){4-4}
      & \textbf{ATE} $\downarrow$ & \textbf{ATE} $\downarrow$ & \textbf{ATE} $\downarrow$ \\
      \midrule
      MASt3R-SLAM \cite{murai2025mast3r} & 0.233 & 0.103 & 0.081 \\
      MUSt3R \cite{cabon2025must3r}      & 0.242 & 0.048 & \cellcolor{red!25}0.046 \\
      CUT3R \cite{wang2025continuous}    & 0.210 & 0.049 & 0.095 \\
      Point3R \cite{wu2025point3r}       & 0.375 & 0.067 & 0.120 \\
      STream3R \cite{lan2025stream3r}    & 0.214 & \cellcolor{orange!25}0.026 & 0.052 \\
      WinT3R \cite{li2025wint3r}                            & 0.225 & 0.074 & 0.062 \\
      StreamVGGT \cite{streamVGGT}       & 0.394 & 0.057 & 0.120 \\
      TTT3R \cite{chen2025ttt3r}         & 0.210 & 0.028 & 0.064 \\
      \midrule
      \textbf{Ours (CUT3R)}    & \cellcolor{red!25}0.168 & 0.033          & 0.092          \\
      \textbf{Ours (STream3R)} & \cellcolor{orange!25}0.171       & \cellcolor{red!25}0.018 & \cellcolor{orange!25}0.049 \\
      \bottomrule
    \end{tabular}
    }
  \end{minipage}
\end{table*}
\subsection{Quantitative Results}

\subsubsection{Camera Pose Estimation.}
We evaluate ATE across multiple benchmarks, including KITTI~\cite{Geiger2012CVPR} and Virtual KITTI (vKITTI)~\cite{cabon2020virtual} for outdoor driving scenarios, as well as Sintel~\cite{Butler:ECCV:2012}, TUM-Dynamic~\cite{sturm2012benchmark}, and ScanNet~\cite{dai2017scannet} for diverse indoor and synthetic environments. Following CUT3R~\cite{wang2025continuous} and STream3R~\cite{lan2025stream3r}, we apply Sim(3) alignment to the ground truth before computing ATE. As shown in \cref{tab:kitti_final_refined,tab:vkitti_final_refined,tab:sintel_tum_scannet_results}, Scal3R consistently outperforms both offline and online baselines. On KITTI (\cref{tab:kitti_final_refined}), our method achieves an average ATE of $69.7$, reducing error by over $60\%$ compared to the strongest online competitor TTT3R ($182.2$), with particularly pronounced gains on long-range sequences such as Seq.~00 and Seq.~02. On vKITTI (\cref{tab:vkitti_final_refined}), Scal3R~(CUT3R) attains an average ATE of $5.63$, surpassing all streaming methods by a large margin and approaching the accuracy of the offline method $\pi^3$, while remaining fully online. On Sintel, TUM-Dynamic, and ScanNet (\cref{tab:sintel_tum_scannet_results}), Scal3R generalizes robustly to unseen domains: Scal3R~(CUT3R) achieves the best ATE of $0.168$ on Sintel, and Scal3R~(STream3R) attains state-of-the-art ATE of $0.018$ on TUM-Dynamic and $0.049$ on ScanNet, demonstrating strong performance across both large-scale outdoor and dense indoor environments without sacrificing online processing.

\subsubsection{3D Reconstruction.}
We evaluate 3D reconstruction quality on the 7-Scenes dataset using 300-frame sequences, reporting Accuracy (Acc.), Completeness (Comp.), and Normal Consistency (NC). As shown in \cref{tab:7scenes_reconstruction}, Scal3R variants consistently enhance the geometric consistency of their frozen backbones. Ours~(STream3R) achieves the best performance across all metrics, attaining an NC mean of $0.579$ and median of $0.622$, surpassing the STream3R backbone by a clear margin. Notably, while competing streaming methods struggle to improve geometric consistency beyond their base backbone, our scale-decoupled formulation yields consistent gains in NC without sacrificing accuracy or completeness.

\begin{table}[t]
\setlength{\tabcolsep}{3pt}
  \caption{\textbf{3D reconstruction on 7-Scenes (300 frames).} We report Accuracy (Acc.), Completeness (Comp.), and Normal Consistency (NC), each with mean and median values. Scal3R consistently improves upon both CUT3R and STream3R backbones, achieving the best NC across all sequences.}
  \label{tab:7scenes_reconstruction}
  \centering
  \small
  \begin{tabular}{l cc cc cc}
    \toprule
    \multirow{3}{*}{\textbf{Methods}} & \multicolumn{6}{c}{\textbf{7-Scenes (300 frames)}} \\
    \cmidrule(lr){2-7}
    & \multicolumn{2}{c}{\textbf{Acc.} $\downarrow$} & \multicolumn{2}{c}{\textbf{Comp.} $\downarrow$} & \multicolumn{2}{c}{\textbf{NC} $\uparrow$} \\
    \cmidrule(lr){2-3} \cmidrule(lr){4-5} \cmidrule(lr){6-7}
    & \textbf{Mean} & \textbf{Med.} & \textbf{Mean} & \textbf{Med.} & \textbf{Mean} & \textbf{Med.} \\
    \midrule
    CUT3R \cite{wang2025continuous} & 0.130 & 0.090 & 0.062 & 0.023 & 0.544 & 0.565 \\
    STream3R \cite{lan2025stream3r} & 0.095 & 0.040 & 0.030 & 0.006 & 0.560 & 0.590 \\
    \midrule
    \textbf{Ours (CUT3R)}    & \cellcolor{orange!25}0.069 & \cellcolor{orange!25}0.034 & \cellcolor{orange!25}0.035 & \cellcolor{orange!25}0.010 & \cellcolor{orange!25}0.566 & \cellcolor{orange!25}0.601 \\
    \textbf{Ours (STream3R)} & \cellcolor{red!25}0.052 & \cellcolor{red!25}0.012 & \cellcolor{red!25}0.020 & \cellcolor{red!25}0.004 & \cellcolor{red!25}0.579 & \cellcolor{red!25}0.622 \\
    \bottomrule
  \end{tabular}
\end{table}

\begin{figure*}[t]
    \centering
    \includegraphics[width=\textwidth]{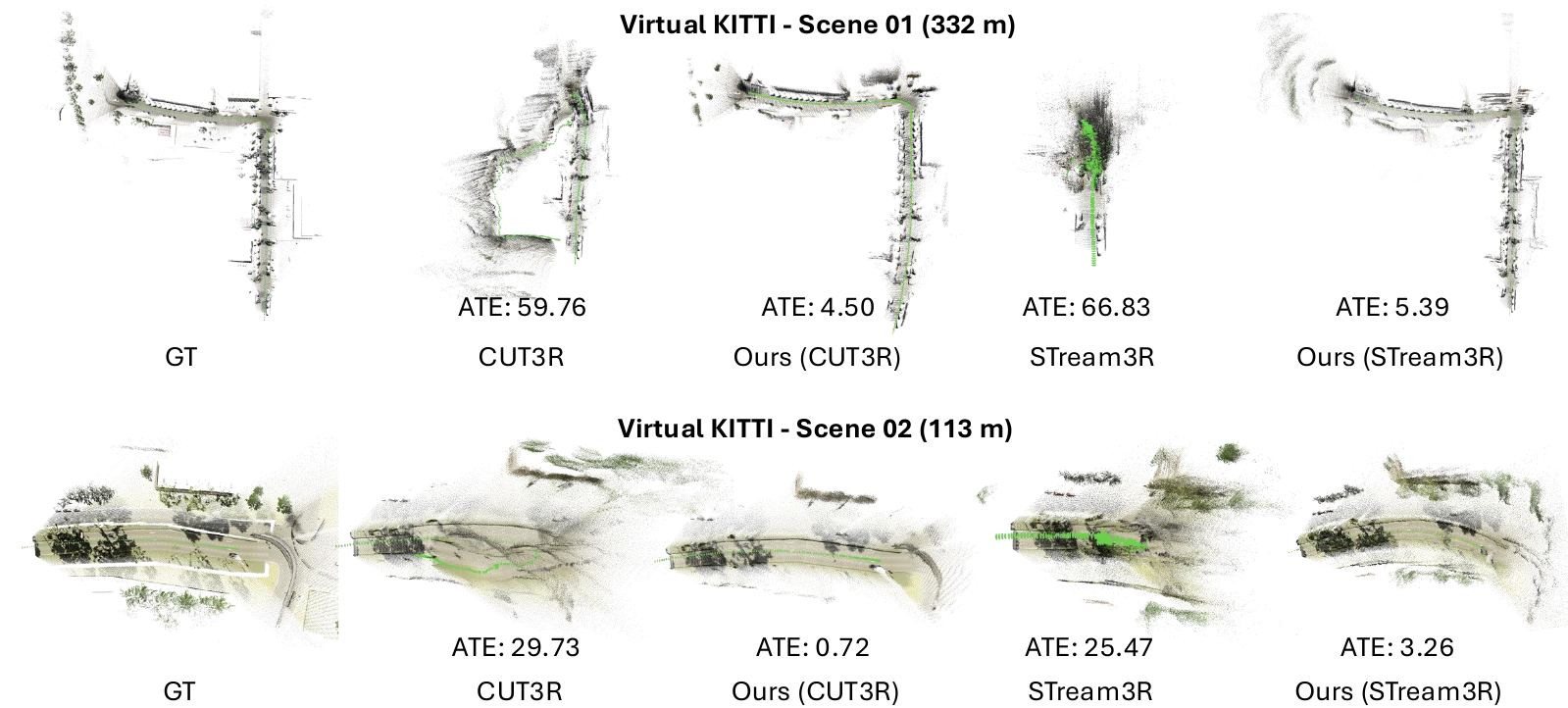}
    \caption{\textbf{Qualitative 3D reconstruction on Virtual KITTI.} Comparison against CUT3R and STream3R on Scene~01 (332\,m) and Scene~02 (113\,m). While both baselines produce distorted or collapsed point clouds, Scal3R recovers scene geometry closely matching the ground truth across both sequences.}
    \label{fig:vkitti_qual}
\end{figure*}

\begin{figure*}[t]
    \centering
    \includegraphics[width=\textwidth]{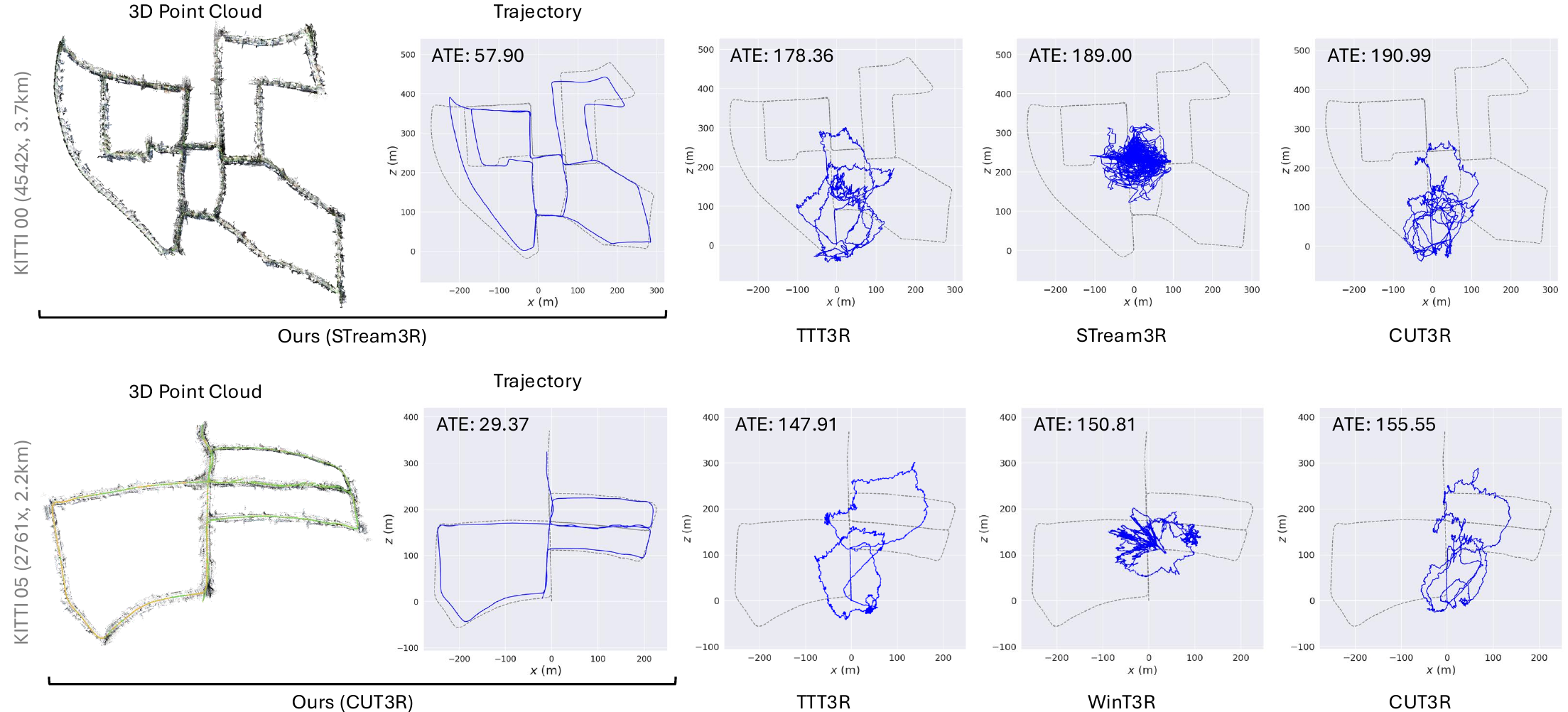}
    \caption{\textbf{Qualitative trajectory comparison on KITTI long sequences.} We visualize estimated camera trajectories on Seq.~00 and Seq.~05 against CUT3R, WinT3R, and STream3R. All baselines suffer from catastrophic drift, while Scal3R faithfully recovers the full loop structure with metric accuracy.}
    \label{fig:pose_qual}
\end{figure*}

\begin{figure*}[t]
    \centering
    \includegraphics[width=\textwidth]{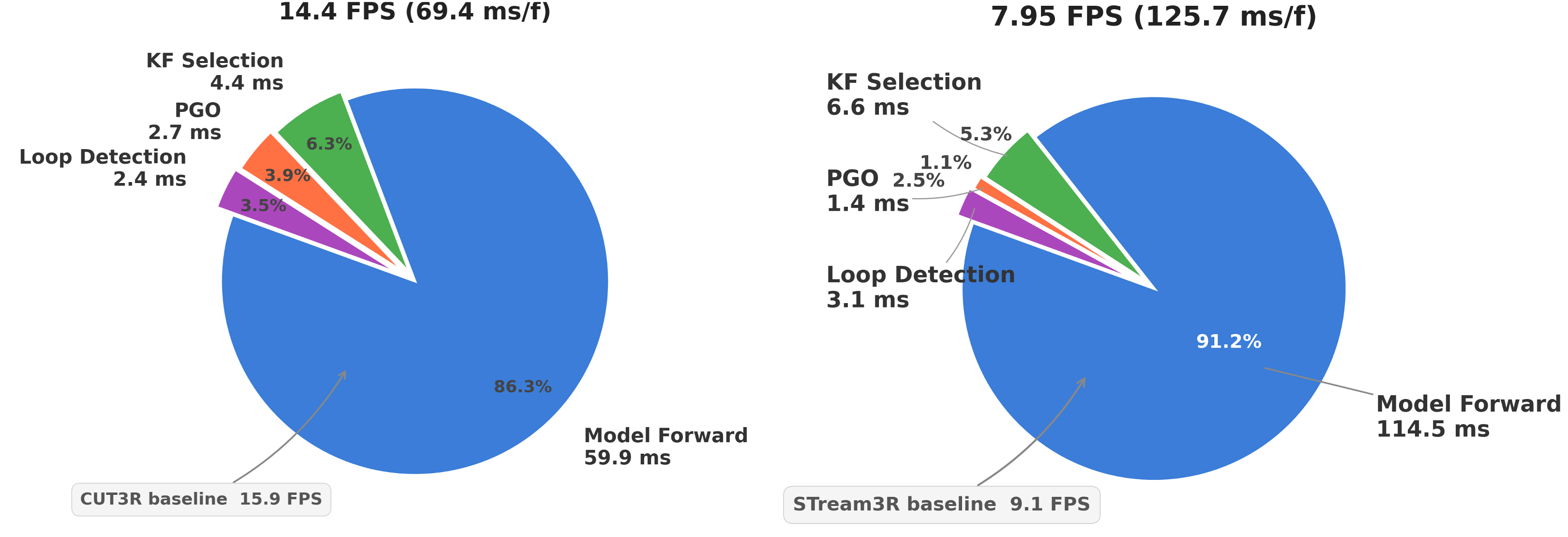}
    \caption{\textbf{Per-frame runtime breakdown on KITTI.} On both the CUT3R (left) and STream3R (right) backbones, the frozen forward pass dominates latency; keyframe selection, PGO, and loop detection together add only a small fraction.}
    \label{fig:runtime}
\end{figure*}

\subsection{Qualitative Results}
\cref{fig:vkitti_qual,fig:pose_qual} visualize 3D reconstruction and trajectory estimation on outdoor long-sequence benchmarks, confirming stable reconstruction and pose prediction across varying spatial extents.
In terms of 3D reconstruction (\cref{fig:vkitti_qual}), we compare our method against CUT3R and STream3R on Virtual KITTI Scene~01 (332\,m) and Scene~02 (113\,m). While CUT3R produces severely distorted point clouds with substantial geometric drift, and STream3R collapses into degenerate reconstructions that deviate considerably from the ground-truth layout, both Ours~(CUT3R) and Ours~(STream3R) recover scene geometry that closely matches the ground truth, with clean structural boundaries and well-preserved spatial extent.
For long-sequence pose estimation (\cref{fig:pose_qual}), we visualize estimated trajectories on KITTI Seq.~00 and Seq.~05 against CUT3R, WinT3R, and STream3R. All three baselines suffer from catastrophic trajectory collapse, producing chaotic, self-intersecting paths that bear no resemblance to the ground-truth loop structure. In contrast, Ours~(CUT3R) faithfully traces the full loop trajectory, maintaining metric accuracy and geometric coherence across hundreds of meters.

\subsection{Ablation Study}
We conduct ablation studies on vKITTI and KITTI to validate the key components of our method, covering model training strategy, inference-time design choices, and loop closure. Results are summarized in \cref{tab:ablation_train,tab:ablation_infer,tab:ablation_loop}.

\subsubsection{Model Training Strategy.}
As shown in \cref{tab:ablation_train}, removing reference-frame supervision entirely (\emph{w/o reference}) sharply degrades RPE\textsubscript{trans} to 3.336, while a single reference reduces ATE to 15.764. Our full multi-reference training achieves the best ATE of 5.632, confirming that denser reference supervision is critical for robust long-sequence pose estimation.

\begin{table*}[t]
  \centering
  \begin{minipage}[t]{0.53\textwidth}
    \centering
    \caption{\textbf{Ablation on training strategy (vKITTI).} We vary the number of reference frames used for supervision during training. Denser reference supervision consistently lowers ATE, while removing references entirely causes severe RPE degradation.}
  \label{tab:ablation_train}
  \centering
\setlength{\tabcolsep}{3pt}
  \resizebox{\textwidth}{!}{
  \begin{tabular}{@{}lccc@{}}
    \toprule
    Method & ATE $\downarrow$ & RPE\textsubscript{trans} $\downarrow$ & RPE\textsubscript{rot} $\downarrow$ \\
    \midrule
    Baseline (CUT3R) & 56.390 & 1.678 & 0.342 \\
    w/o reference   & 33.318 & 3.336 & 0.842 \\
    1 reference     & 15.764 & 0.217 & 0.448 \\
    \midrule
    Ours            & \textbf{5.632} & \textbf{0.177} & \textbf{0.485} \\
    \bottomrule
  \end{tabular}
  }
  \end{minipage}
  \hfill
  \begin{minipage}[t]{0.45\textwidth}
    \centering
    \caption{\textbf{Ablation on inference-time components (vKITTI).} Both keyframe selection and PGO are essential, and scaling the reference count from 4 to 12 at inference time consistently reduces ATE without retraining.}
  \label{tab:ablation_infer}
  \centering
\setlength{\tabcolsep}{3pt}
  \resizebox{\textwidth}{!}{
  \begin{tabular}{@{}lccc@{}}
    \toprule
    Method & ATE $\downarrow$ & RPE\textsubscript{trans} $\downarrow$ & RPE\textsubscript{rot} $\downarrow$ \\
    \midrule
    w/o keyframe    & 38.258 & 0.787 & 1.388 \\
    w/o PGO         & 20.089 & 0.203 & 0.544 \\
    $K=4$           & 15.748 & 0.188 & \textbf{0.470} \\
    $K=8$           &  7.362 & 0.180 & \textbf{0.470} \\
    \midrule
    Full ($K=12$)   & \textbf{5.632} & \textbf{0.177} & 0.485 \\
    \bottomrule
  \end{tabular}
  }
  \end{minipage}
\end{table*}

\subsubsection{Inference-Time Components.}
\cref{tab:ablation_infer} ablates the inference-time pipeline. Removing keyframe selection or PGO each leaves a large gap to the full system (ATE: 38.258 without keyframe selection). Progressively increasing reference count from 4 to 12 then consistently reduces ATE from 15.748 to 5.632.


\subsubsection{Runtime Analysis.}
\cref{fig:runtime} breaks down the latency on KITTI ($K{=}12$). The frozen forward pass dominates on both backbones (86.3\% on CUT3R, 91.2\% on STream3R), so keyframe selection, PGO, and loop detection add little: the full pipeline runs at 14.4 and 7.95\,FPS, versus 15.9 and 9.1\,FPS for the backbones.

\subsubsection{Loop Closure.}
As shown in \cref{tab:ablation_loop} and \cref{fig:ablation_loop}, loop closure reduces average ATE on KITTI from 143.45 to 75.01 (48\% improvement), with particularly large gains on sequences containing large loops (\eg, Seq.~00 and Seq.~05), where globally consistent trajectory correction is most needed.

\subsubsection{Robustness Analysis.}
\begin{figure*}[t]
    \centering
    \begin{minipage}[t]{0.48\textwidth}
        \centering
        \includegraphics[width=\textwidth]{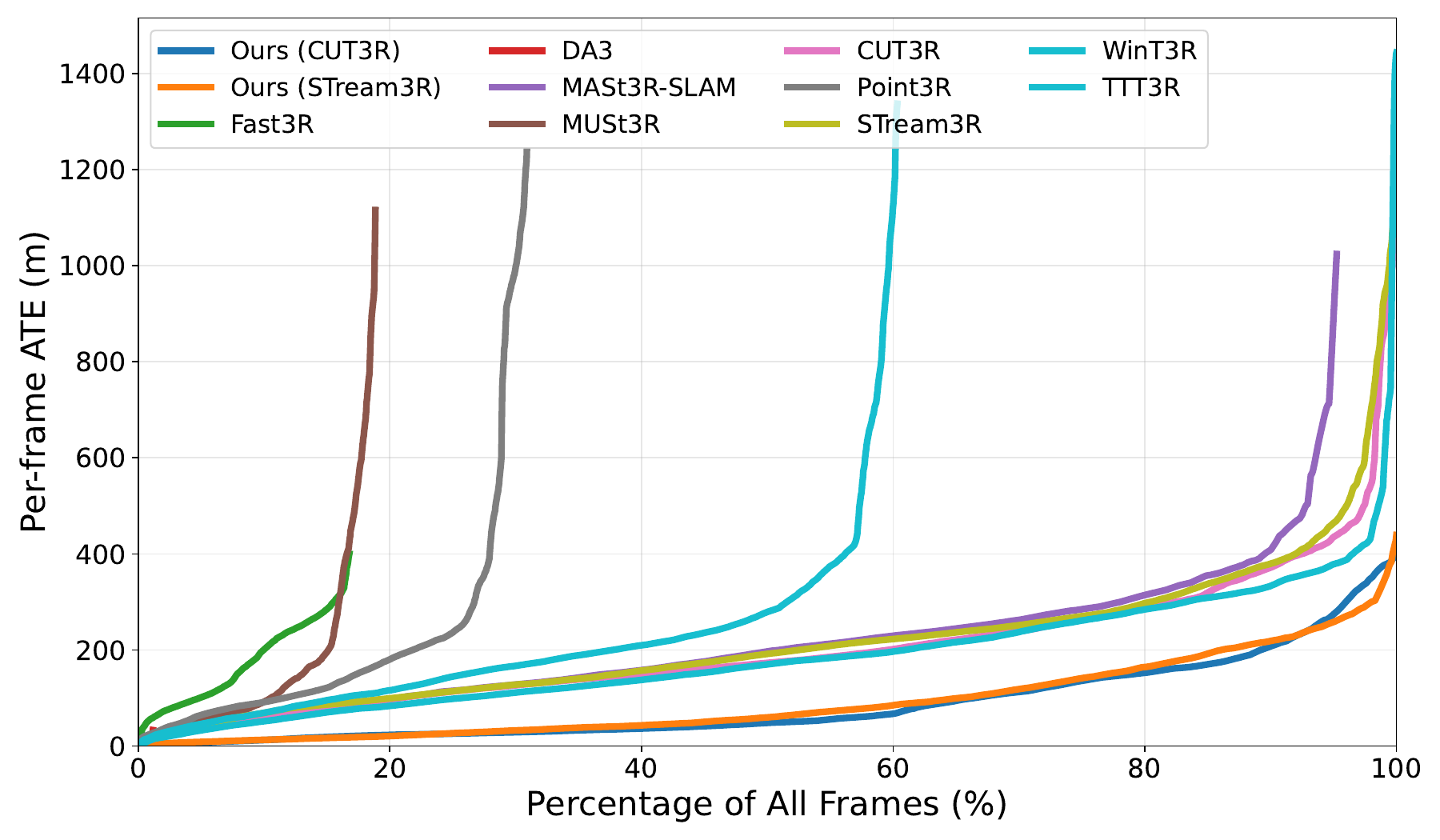}
        \caption{\textbf{Robustness Analysis on KITTI.} Scal3R maintains the lowest error across the full evaluation range, while competing methods suffer catastrophic divergence at moderate sequence lengths.}
        \label{fig:rebust_analysis}
    \end{minipage}
    \hfill
    \begin{minipage}[t]{0.48\textwidth}
        \centering
        \vspace{-30mm}
        \includegraphics[width=\textwidth]{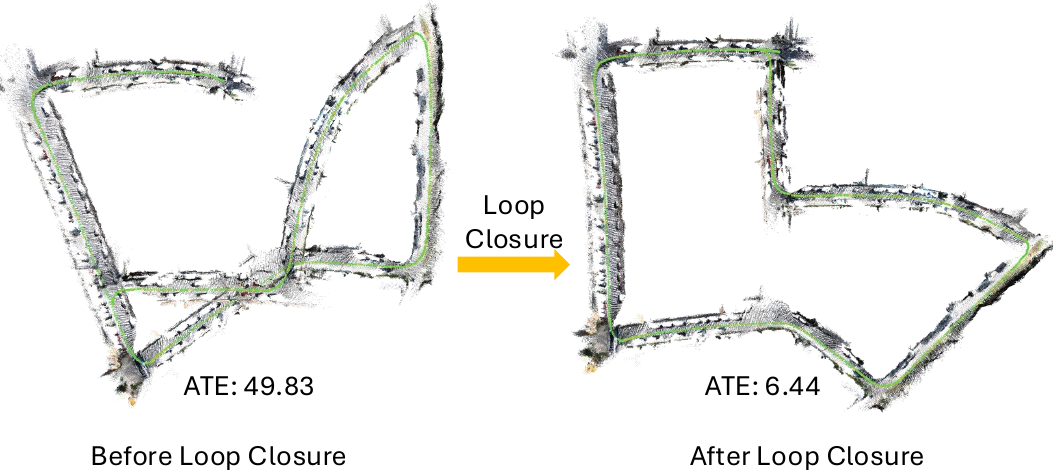}
        \caption{\textbf{Qualitative effect of loop closure on KITTI 07.} Without loop closure, accumulated drift causes the trajectory to deviate from the ground-truth loop structure. With loop closure, Scal3R recovers a globally consistent trajectory.}
        \label{fig:ablation_loop}
    \end{minipage}
\end{figure*}

\cref{fig:rebust_analysis} presents per-frame ATE curves sorted in ascending order across all KITTI sequences. Methods such as MUSt3R and Point3R suffer catastrophic failures at moderate sequence lengths, while our method maintains the lowest per-frame ATE throughout the entire evaluation range, confirming superior robustness under challenging long-sequence conditions.

\begin{table}[t]
\setlength{\tabcolsep}{3pt}
  \caption{\textbf{Ablation on loop closure (KITTI, ATE $\downarrow$).} Loop closure reduces average ATE by 48\%, with the largest gains on loop-heavy sequences (\eg, Seq.~00 and 05).}
  \label{tab:ablation_loop}
  \centering
  \small
  \resizebox{\textwidth}{!}{
  \begin{tabular}{@{}llccccccc|c@{}}
    \toprule
    \multirow{2}{*}{Method} & \multirow{2}{*}{LC} & \multicolumn{7}{c|}{KITTI (ATE $\downarrow$)} & \multirow{2}{*}{Avg.} \\
    \cmidrule(lr){3-9}
    & & 00 & 02 & 05 & 06 & 07 & 08 & 09 & \\
    \midrule
    \multirow{2}{*}{Ours (CUT3R)}
      & & 185.70 & 235.28 & 108.73 & 84.86 & 49.83 & 241.86 & 97.86 & 143.45 \\
      & \checkmark & \cellcolor{red!25}45.34 & \cellcolor{red!25}139.88 & \cellcolor{red!25}29.37 & \cellcolor{orange!25}47.40 & \cellcolor{red!25}6.44 & \cellcolor{orange!25}227.02 & \cellcolor{red!25}29.63 & \cellcolor{red!25}75.01 \\
    \midrule
    \multirow{2}{*}{Ours (STream3R)}
      & & 166.37 & 239.36 & 134.08 & 35.95 & 60.68 & 195.20 & 74.18 & 129.40 \\
      & \checkmark & \cellcolor{orange!25}57.90 & \cellcolor{orange!25}170.75 & \cellcolor{orange!25}39.05 & \cellcolor{red!25}18.07 & \cellcolor{orange!25}15.19 & \cellcolor{red!25}173.83 & \cellcolor{orange!25}73.91 & \cellcolor{orange!25}78.39 \\
    \bottomrule
  \end{tabular}
  }
\end{table}

\section{Conclusion}
We presented Scal3R, which tackles the instability of global pose regression on long sequences by reformulating camera localization as multi-reference relative pose querying on a frozen backbone. Lightweight tokens ($\sim$1\% of parameters) query relative poses that an online pose-graph backend aggregates into a globally consistent trajectory. Trained in 8 hours on a single GPU, Scal3R enables accurate online reconstruction of long video streams.

\subsubsection{Limitations.}
First, performance is bounded by the frozen backbone, degrading when it fails under occlusion or textureless regions. Second, the online backend has its own weaknesses: appearance-based loop closure can miss revisits under extreme viewpoint or illumination change, and keyframe selection and loop detection rely on hand-set thresholds. Improving both remains future work.

\vspace{-.5em}
\subsubsection*{Acknowledgements.}
This work was supported by NVIDIA Taiwan AI Research \& Development Center (TRDC). This research was funded by the National Science and Technology Council, Taiwan, under Grants NSTC 112-2222-E-A49-004-MY2, 113-2628-E-A49-023-, 115-2628-E-A49-024-, and 111-2628-E-A49-018-MY4. Yu-Lun Liu acknowledges the Yushan Young Fellow Program by the MOE in Taiwan.



%
%
\bibliographystyle{splncs04}
\bibliography{main}

\renewcommand{\thesection}{\Alph{section}}
\renewcommand{\thesubsection}{\Alph{section}.\arabic{subsection}}
\renewcommand{\thesubsubsection}{\Alph{section}.\arabic{subsection}.\arabic{subsubsection}}

\appendix
\clearpage
\section*{Overview}
This supplementary material provides additional details and experiments that complement the main paper.
\cref{sec:supp-impl} describes implementation details, including the inference pipeline algorithm, evaluation protocol, and PGO parameters.
\cref{sec:supp-vis} presents additional qualitative results and ablation visualizations on vKITTI, KITTI, and TUM-Dynamic.
\cref{sec:supp-profile} reports runtime and memory profiling, including a per-component latency breakdown and scalability analysis with respect to the reference count $K$.
\cref{sec:supp-exp} provides additional experiments: metric-scale pose estimation (\cref{sec:supp-metric}), comparison with classic SLAM systems (\cref{sec:supp-slam}), a comparison with the concurrent LongStream (\cref{sec:supp-longstream}), an ablation on asymmetric vs.\ symmetric attention injection (\cref{sec:supp-attn-ablation}), compatibility with zero-shot test-time training (\cref{sec:supp-zeroshot}), and robustness analysis on dynamic scenes (\cref{sec:supp-dynamic}).

\section{Implementation Details}
\label{sec:supp-impl}

We implement Scal3R using PyTorch, building upon the frozen CUT3R and STream3R backbones without modifying their pretrained weights.
Pose-graph optimization is performed using GTSAM's iSAM2 incremental solver, with a gap-dependent noise model $\sigma = \sigma_{\text{base}} \cdot \Delta^{0.5}$ and Huber robust kernel for outlier rejection.
Loop closure retrieval employs a pretrained DINOv2-B backbone with a SALAD aggregation layer, indexed online via FAISS over keyframe descriptors.
All models are trained on TartanAir with the AdamW optimizer at a learning rate of $1 \times 10^{-4}$, batch size 8, for 40 epochs.
Training converges in approximately 8 hours on a single NVIDIA A100 GPU.
All inference experiments are evaluated on NVIDIA A100 GPUs.

\subsection{Inference Pipeline}
\label{sec:supp-algo}

\Cref{alg:inference} summarizes the complete per-frame inference pipeline of Scal3R.
At each timestep, reference frames are first selected from the pose token buffer,
followed by loop closure detection over archived keyframe descriptors via DINOv2-SALAD indexed with FAISS.
If a loop candidate is detected, its archived camera token $\mathbf{c}_{r_{\text{loop}}}$ is re-injected into the pose token buffer as an additional reference slot before model inference,
requiring no architectural modification.
The frozen backbone then predicts both the local pointmap $X_t$ and multi-reference relative poses $\{\hat{\mathbf{T}}_{t \leftarrow r_k}\}$ via asymmetric attention injection.
These pairwise constraints, together with any loop closure edge, are registered into the factor graph and incrementally optimized via iSAM2,
with corrected poses immediately written back to the pose token buffer to benefit subsequent frames.
Finally, keyframe selection determines whether the current frame updates the buffer and the keyframe archive, and buffer pruning maintains a bounded memory footprint throughout the stream.

\begin{algorithm}
\caption{Scal3R Online Inference Pipeline}
\label{alg:inference}
\begin{algorithmic}[1]
\Require Image stream $\{I_1, I_2, \ldots, I_T\}$; frozen backbone; thresholds $\tau_{\text{overlap}}, \tau_{\text{conf}}, \tau_{\text{sim}}, \tau_{\text{gap}}, w_{\text{nms}}$
\Ensure Globally consistent camera poses $\{T_1, \ldots, T_T\} \subset SE(3)$
\State Initialize pose token buffer $\mathcal{B} \leftarrow \emptyset$, keyframe archive $\mathcal{A} \leftarrow \emptyset$, factor graph $\mathcal{G} \leftarrow \emptyset$
\For{each frame $I_t$}

  \State \textbf{// Reference Selection}
  \State Retrieve reference indices $\{r_1, \ldots, r_K\}$ from $\mathcal{B}$

  \State \textbf{// Loop Closure Detection}
  \State $d_t \leftarrow \text{DINOv2-SALAD}(I_t)$; query FAISS over $\mathcal{A}$
  \If{$\exists\, r_{\text{loop}}$: similarity $> \tau_{\text{sim}}$, gap $> \tau_{\text{gap}}$, NMS window $w_{\text{nms}}$}
    \State Inject archived camera token $\mathbf{c}_{r_{\text{loop}}}$ from $\mathcal{A}$ into $\mathcal{B}$ as additional reference slot
  \EndIf

  \State \textbf{// Model Inference}
  \State $F_t \leftarrow \text{Encoder}(I_t)$ \Comment{frozen ViT}
  \For{$k = 1, \ldots, K$}
    \State $\tilde{\mathbf{q}}_k \leftarrow \mathbf{q} + \mathrm{MLP}(\mathbf{c}_{r_k})$ \Comment{pose token assembly, Eq.~(1)}
  \EndFor
  \State $\{X_t,\, \tilde{\mathbf{q}}_k^L\} \leftarrow \text{Decoder}(F_t,\, \{\tilde{\mathbf{q}}_k\},\, \text{state})$ \Comment{asymmetric attention, Eqs.~(2)--(3)}
  \State $\hat{\mathbf{T}}_{t \leftarrow r_k} \leftarrow \mathrm{Head}(\tilde{\mathbf{q}}_k^L) \in SE(3)$ for all $k$ \Comment{Eq.~(4)}

  \State \textbf{// Pose-Graph Optimization}
  \State $T_t \leftarrow T_{r_1} \cdot \hat{\mathbf{T}}_{t \leftarrow r_1}^{-1}$ \Comment{chain initialization}
  \For{$k = 1, \ldots, K$}
    \State Add between-factor $(\hat{\mathbf{T}}_{t \leftarrow r_k},\, \Sigma_{t,k})$ to $\mathcal{G}$ \Comment{Eq.~(6); $\sigma{=}\sigma_{\text{base}}{\cdot}\Delta^{0.5}$}
  \EndFor
  \If{loop edge exists}
    \State Add loop edge $(\hat{\mathbf{T}}_{t \leftarrow r_{\text{loop}}},\, \Sigma_{\text{loop}})$ to $\mathcal{G}$ with tight Gaussian noise
  \EndIf
  \State $\{T_i\} \leftarrow \text{iSAM2.optimize}(\mathcal{G})$; write back optimized poses to $\mathcal{B}$

  \State \textbf{// Keyframe Selection \& Buffer Update}
  \State Compute overlap score via KD-tree on $X_t$
  \If{overlap score $< \tau_{\text{overlap}}$ \textbf{and} median conf $> \tau_{\text{conf}}$}
    \State Update KV cache; append $\mathbf{c}_t$ to $\mathcal{B}$; archive $\mathbf{c}_t$ in $\mathcal{A}$
  \Else
    \State Discard KV state snapshot; do not update $\mathcal{B}$
  \EndIf
  \State Retain most recent $K_{\text{kf}}$ keyframe and $K_{\text{nkf}}$ non-keyframe entries in $\mathcal{B}$

\EndFor
\State \Return $\{T_t\}$ from final iSAM2 marginalization
\end{algorithmic}
\end{algorithm}

\subsection{Evaluation Protocol and Scale Handling}
\label{sec:supp-scale}

Since Scal3R freezes the pretrained backbone and supervises only relative pose in a scale-normalized space,
the predicted trajectory operates in the backbone's internal scale rather than metric scale.
During training, both the predicted and ground-truth point clouds are independently normalized by their respective scale factors,
and a robust scale alignment is applied before loss computation to eliminate monocular scale ambiguity.
At inference, relative poses are chained and refined via iSAM2 entirely within this model-internal scale space.

For evaluation, we follow the protocol of CUT3R~\cite{wang2025continuous} and STream3R~\cite{lan2025stream3r},
applying Sim(3) alignment via the Umeyama method to align the predicted trajectory to the ground truth before computing ATE.
This alignment recovers the global scale, rotation, and translation, ensuring fair comparison across all methods regardless of their internal scale convention.

Since the CUT3R backbone is trained with metric-scale supervision, it retains absolute scale information in its camera tokens.
We therefore additionally evaluate Scal3R in a metric-scale setting, where the Sim(3) scale factor is fixed to 1 (SE(3) alignment),
and report these results in \cref{sec:supp-metric}.

\subsection{PGO Parameters}
\label{sec:supp-pgo}

\subsubsection{Keyframe Selection.}
Frames are designated as keyframes when their depth-normalized 3D overlap score falls below $\tau_{\text{overlap}} = 0.1$ and median depth confidence exceeds $\tau_{\text{conf}} = 1.2$.
The first $N_{\text{init}} = 5$ frames are unconditionally treated as keyframes to initialize the system.
The active pose token buffer retains the most recent $K_{\text{kf}} = 4$ to $12$ keyframes depending on the dataset, with additional non-keyframe slots enabled for outdoor driving sequences.
For outdoor driving sequences (KITTI, vKITTI), the frozen decoder's streaming state is reset every $N_{\text{reset}} = 10$ keyframes to prevent memory overflow and feature degradation; for all other datasets the state is never reset.

\subsubsection{PGO Noise Model.}
We set $\sigma_{\text{base}} = 0.5$ for both rotation and translation, with gap-dependent scaling $\sigma = \sigma_{\text{base}} \cdot \Delta^{0.5}$ as described in the main paper.
Sequential edges are weighted by a Huber robust kernel ($k = 1.345$) to downweight outlier constraints, while loop closure edges omit the robust kernel to enforce tight trajectory correction.
Pose-graph optimization is performed via iSAM2 with a relinearization threshold of $0.1$.

\subsubsection{Loop Closure Detection.}
Candidates are filtered by cosine similarity threshold $\tau_{\text{sim}} = 0.85$, minimum temporal gap $\tau_{\text{gap}} = 200$ frames, and non-maximum suppression within a window of $w_{\text{nms}} = 50$ frames.
At most one loop edge is injected per keyframe.

\section{Additional Visualizations}
\label{sec:supp-vis}

\subsection{More Qualitative Results}
\label{sec:supp-pts}

\begin{figure*}[t]
    \centering
    \includegraphics[width=\textwidth]{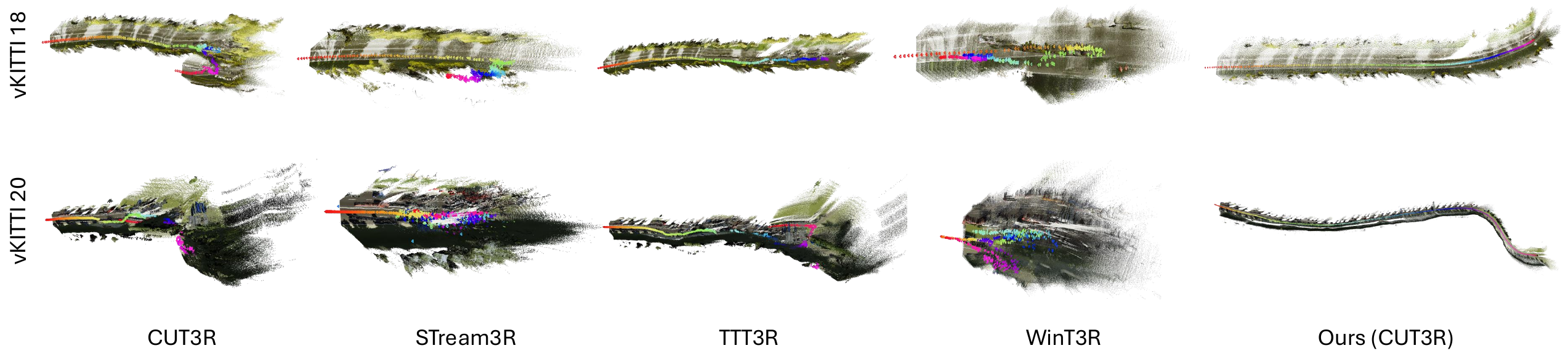}
    \vspace{-6mm}
    \caption{
    \textbf{More qualitative comparison on vKITTI.}
    Scal3R produces globally consistent reconstructions with minimal trajectory drift, while baselines exhibit elongated or distorted point clouds.
    }
    \label{fig:more_qual}
\end{figure*}

\begin{figure*}[t]
    \centering
    \includegraphics[width=\textwidth]{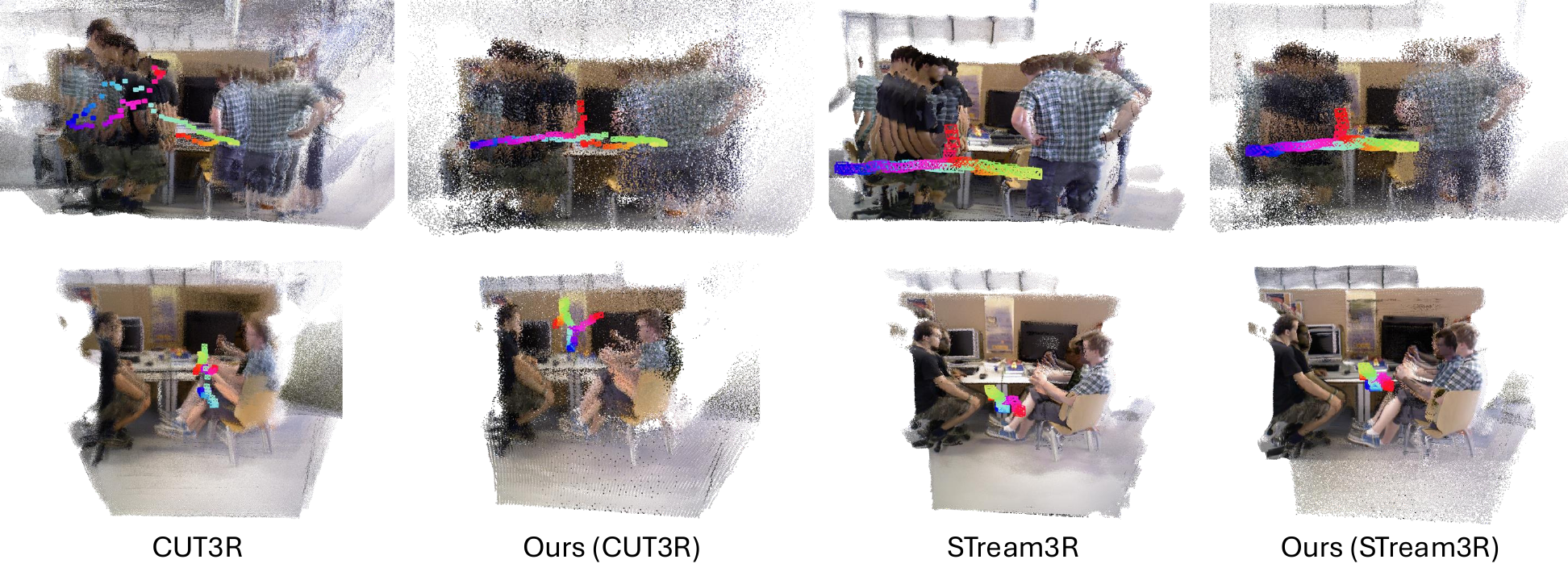}
    \vspace{-6mm}
    \caption{
    \textbf{More qualitative comparison on TUM-Dynamic.}
    Scal3R produces cleaner reconstructions with fewer ghosting artifacts from dynamic pedestrians on both backbones.
    }
    \label{fig:more_qual_tum}
\end{figure*}

\cref{fig:more_qual} presents additional qualitative comparisons on vKITTI sequences 18 and 20.
CUT3R and STream3R both exhibit noticeable trajectory drift on these long sequences, producing elongated and skewed reconstructions.
TTT3R maintains reasonable local geometry but accumulates global error, while WinT3R shows severe structural distortion, particularly on seq.\ 20.
In contrast, Scal3R (CUT3R backbone) recovers compact, globally consistent point clouds with well-aligned trajectory shapes on both sequences, benefiting from multi-reference relative pose querying and PGO that jointly suppress cumulative drift over hundreds of frames.
\cref{fig:more_qual_tum} further compares reconstructions on TUM-Dynamic, where moving pedestrians heavily occlude the static scene.
Both CUT3R and STream3R baselines produce fragmented point clouds with ghosting artifacts from the dynamic persons, and their trajectories scatter erratically.
Our variants on both backbones yield cleaner reconstructions with sharper room geometry and more coherent camera trajectories, consistent with the implicit dynamic-object down-weighting observed in the attention maps (\cref{sec:supp-dynamic}).

\subsection{Ablation Visualizations}
\label{sec:supp-pts-ablation}

\begin{figure*}[t]
    \centering
    \includegraphics[width=\textwidth]{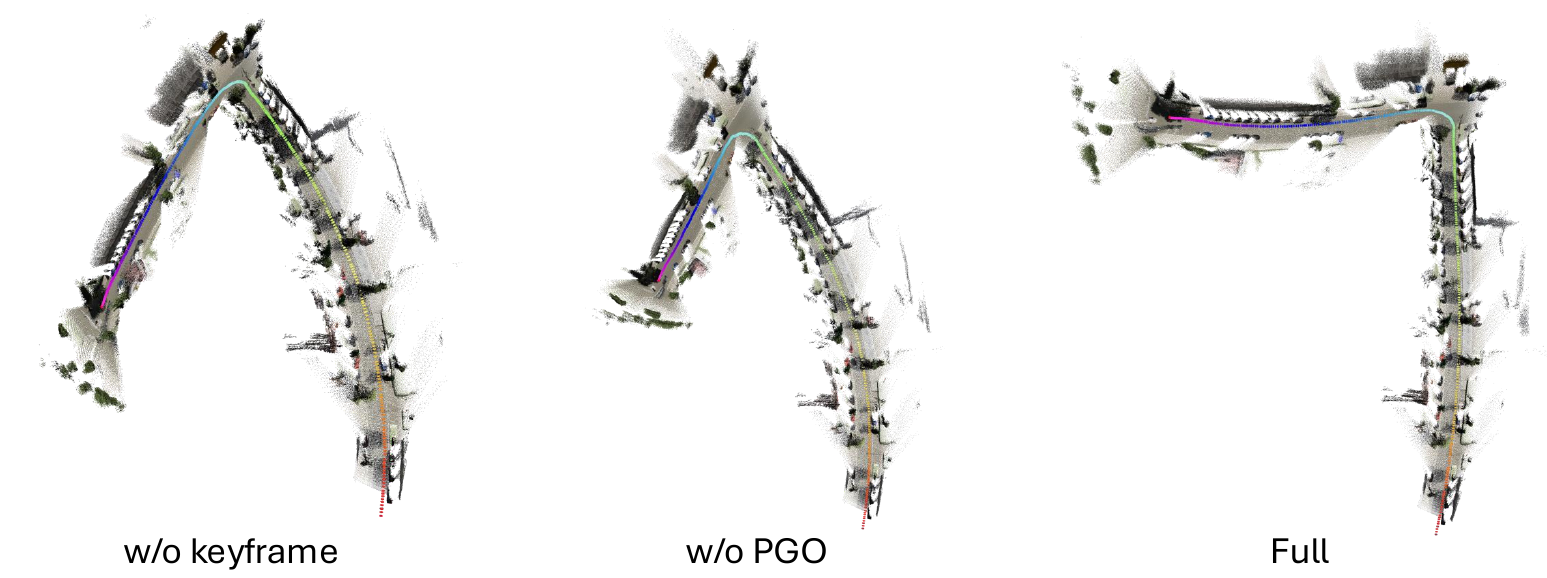}
    \vspace{-6mm}
    \caption{
    \textbf{Ablation visualizations on vKITTI.}
    Removing keyframe selection or PGO each leads to distinct trajectory degradation.
    }
    \label{fig:ablation_vis}
\end{figure*}

\begin{figure*}[t]
    \centering
    \includegraphics[width=\textwidth]{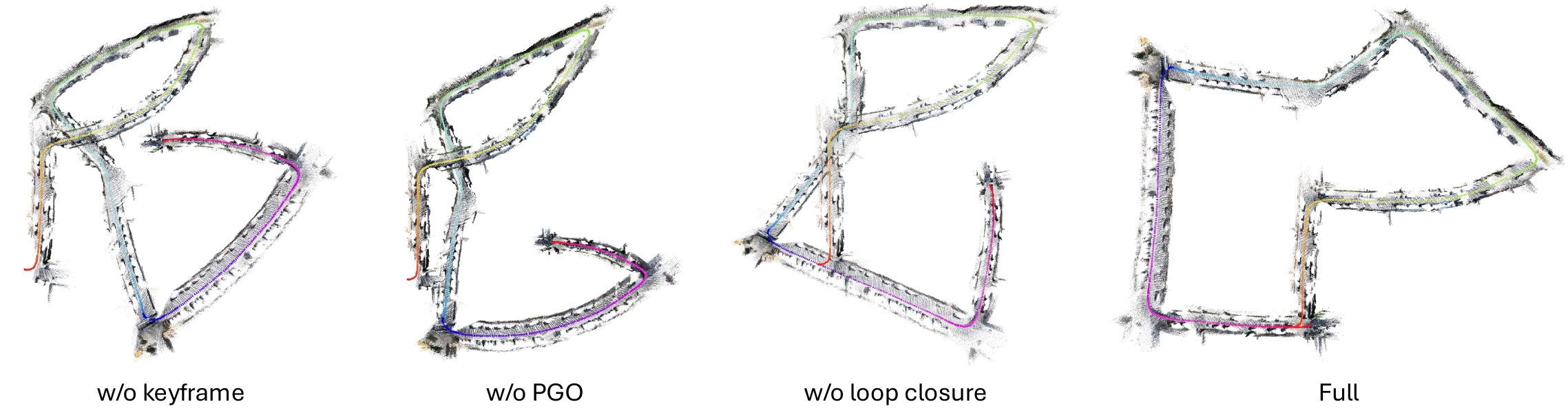}
    \vspace{-6mm}
    \caption{
    \textbf{Ablation visualizations on KITTI.}
    Disabling loop closure leaves visible gaps at revisited regions; the full system produces a globally consistent reconstruction.
    }
    \label{fig:ablation_vis_kitti}
\end{figure*}

\cref{fig:ablation_vis,fig:ablation_vis_kitti} visualize the reconstructed point clouds and estimated trajectories under each ablation setting.
Without keyframe selection, the reference pool lacks geometric diversity, causing severe trajectory drift and distorted global structure on both vKITTI and KITTI.
Removing PGO preserves local smoothness but allows cumulative error to bend the overall trajectory, most visible in the curved segments of vKITTI.
On KITTI (\cref{fig:ablation_vis_kitti}), disabling loop closure leaves a noticeable gap where revisited regions should align, whereas the full system closes the loop and produces a globally consistent reconstruction.

\section{Runtime and Memory Profiling}
\label{sec:supp-profile}

\subsection{Runtime Analysis}
\label{sec:supp-runtime}

\begin{figure*}[t]
    \centering
    \includegraphics[width=\textwidth]{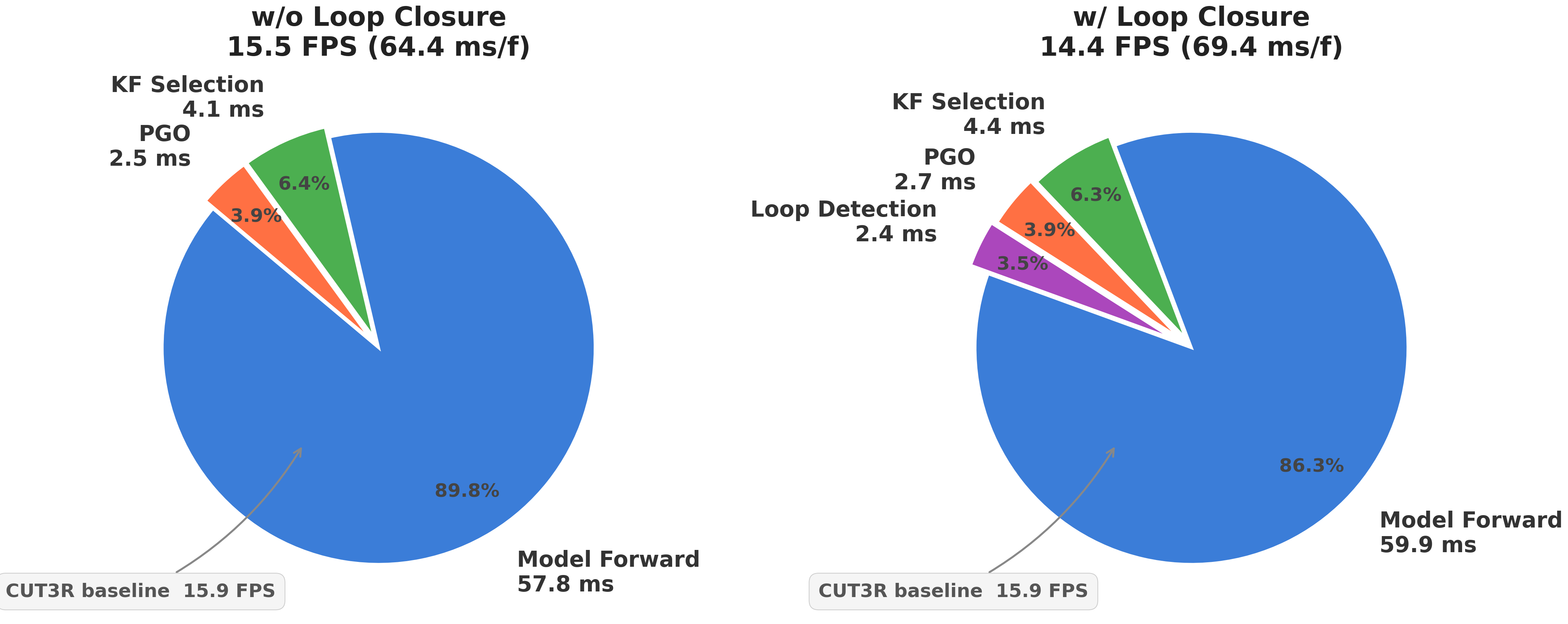}
    \vspace{-6mm}
    \caption{
    \textbf{Scal3R (CUT3R) runtime breakdown on KITTI.}
    The model forward pass dominates latency in both configurations; all system-level components together add less than 10\% overhead.
    }
    \label{fig:runtime}
\end{figure*}

\begin{figure*}[t]
    \centering
    \includegraphics[width=\textwidth]{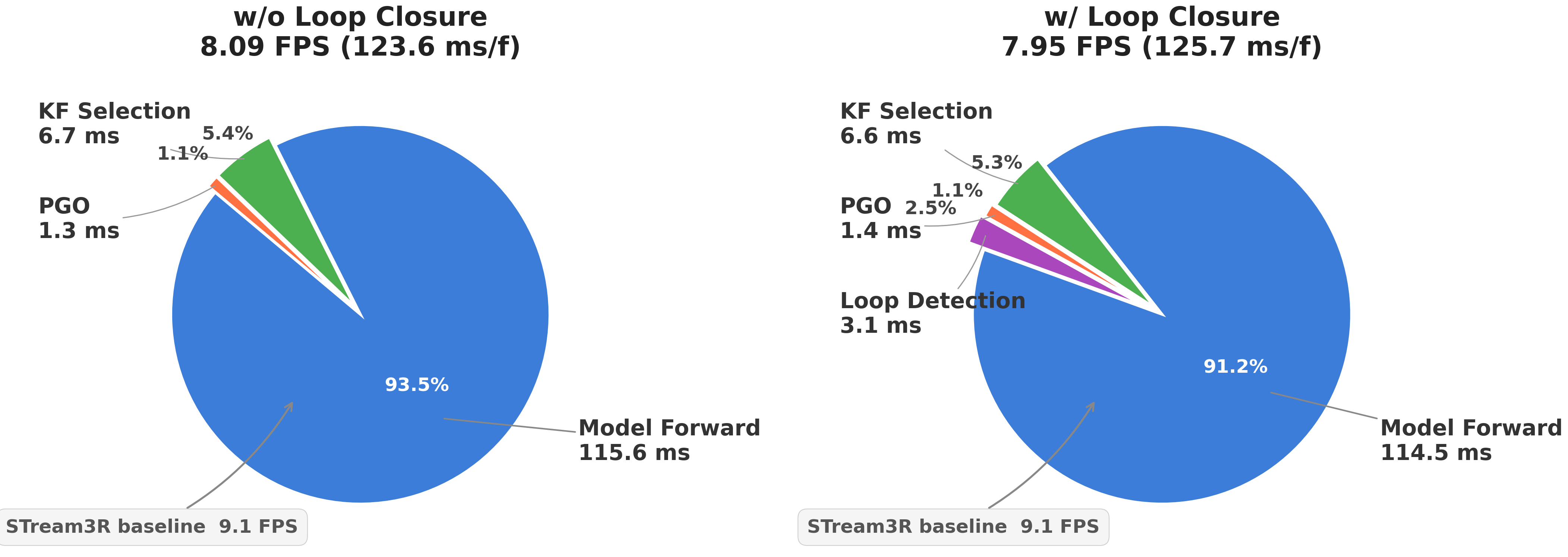}
    \vspace{-6mm}
    \caption{
    \textbf{Scal3R (STream3R) runtime breakdown on KITTI.}
    The latency profile closely mirrors the CUT3R variant; system-level components remain lightweight across both backbones.
    }
    \label{fig:runtime_stream3r}
\end{figure*}

\cref{fig:runtime,fig:runtime_stream3r} break down the per-frame latency of Scal3R on KITTI with $K{=}12$ for both backbones.
On the CUT3R backbone (\cref{fig:runtime}), the model forward pass dominates in both configurations, consuming 89.8\% (57.8\,ms) without loop closure and 86.3\% (59.9\,ms) with it.
Keyframe selection and PGO together add fewer than 7\,ms, while the loop detection module introduces only 2.4\,ms of additional overhead.
Enabling loop closure reduces throughput from 15.5 to 14.4\,FPS, a modest 7\% drop that is justified by the substantial ATE improvements on revisited sequences (cf.\ main paper).
Compared to the CUT3R baseline (15.9\,FPS), the full pipeline retains over 90\% of its throughput.
On the STream3R backbone (\cref{fig:runtime_stream3r}), the latency profile is similar: the forward pass remains the dominant cost, and all system-level components together add less than 10\% overhead.
Across both backbones, the results confirm that keyframe selection, PGO, and loop closure are lightweight relative to the frozen backbone inference, preserving real-time throughput regardless of the backbone choice.

\subsection{Scalability with Reference Count $K$}
\label{sec:supp-K-scale}

\begin{table}[tb]
\centering
\caption{\textbf{Scalability with reference count $K$ on vKITTI.}
ATE is minimized at $K{=}12$; latency grows modestly with $K$.}
\label{tab:scalability-K}
\setlength{\tabcolsep}{5pt}
\begin{tabular}{crrrr}
\toprule
$K$ & ATE $\downarrow$ & FPS $\uparrow$ & ms/frame & Mem (GB) \\
\midrule
4  & 15.75 & 15.84 & 63.23 & 2.91 \\
7  &  7.02 & 15.57 & 64.23 & 2.90 \\
12 &  \textbf{5.63} & 15.10 & 66.23 & 2.94 \\
16 &  8.27 & 14.95 & 66.93 & 3.01 \\
24 & 12.16 & 14.29 & 70.00 & 3.24 \\
\bottomrule
\end{tabular}
\end{table}

\cref{tab:scalability-K} reports ATE, throughput, and memory as a function of the reference count $K$ on vKITTI.
Increasing $K$ from 4 to 12 reduces ATE from 15.75 to 5.63, as additional references provide broader geometric coverage for relative pose querying.
Beyond $K{=}12$, performance degrades ($K{=}24$ yields 12.16 ATE), likely because the pose query tokens are trained with $n{=}3$ references and generalize best within a moderate range.
Meanwhile, latency grows only modestly (63.23\,ms to 70.00\,ms per frame) and peak memory remains below 3.3\,GB across all settings, confirming that asymmetric attention injection scales efficiently with $K$.
We therefore adopt $K{=}12$ as the default at inference, balancing accuracy and throughput at approximately 15\,FPS.

\section{Additional Experiments}
\label{sec:supp-exp}

\subsection{Metric-Scale Pose Estimation}
\label{sec:supp-metric}

\begin{table}[tb]
\centering
\caption{\textbf{Ablation on scale alignment (ATE $\downarrow$).}
SA = scale alignment via Sim(3). Without SA, outdoor ATE degrades substantially for Scal3R, while CUT3R's dominant error remains trajectory drift. We report only the CUT3R backbone variant, as it is trained with metric-scale supervision and thus retains meaningful absolute scale information; the STream3R backbone lacks metric-scale training and is therefore not applicable to this analysis.}
\label{tab:scale-align}
\setlength{\tabcolsep}{4pt}
\begin{tabular}{lcrrrrr}
\toprule
\multirow{2}{*}{Method} & \multirow{2}{*}{SA} & \multicolumn{3}{c}{Indoor} & \multicolumn{2}{c}{Outdoor} \\
\cmidrule(lr){3-5} \cmidrule(lr){6-7}
 & & Sintel & TUM & ScanNet & vKITTI & KITTI \\
\midrule
\multirow{2}{*}{CUT3R}
 & \checkmark & 0.210 & 0.049 & 0.095 & 56.39 & 207.20 \\
 & $\times$   & 0.424 & 0.070 & 0.123 & 68.61 & 216.30 \\
\midrule
\multirow{2}{*}{Scal3R}
 & \checkmark & \textbf{0.168} & \textbf{0.033} & \textbf{0.092} & \textbf{5.63} & \textbf{69.73} \\
 & $\times$   & \textbf{0.381} & \textbf{0.046} & \textbf{0.121} & \textbf{9.25} & \textbf{88.66} \\
\bottomrule
\end{tabular}
\end{table}

\cref{tab:scale-align} ablates the effect of Sim(3) scale alignment on ATE across indoor and outdoor benchmarks.
Without scale alignment, both CUT3R and Scal3R show moderate degradation on indoor scenes (\eg TUM rises from 0.033 to 0.046 for ours), indicating that the predicted poses already carry a reasonable metric scale in small environments.
On outdoor datasets the gap widens substantially: Scal3R degrades from 5.63 to 9.25 on vKITTI and from 69.73 to 88.66 on KITTI, reflecting the difficulty of maintaining consistent scale over kilometer-scale trajectories.
Notably, removing SA changes CUT3R far less on vKITTI (56.39 to 68.61) than its already large drift, suggesting its dominant error source is trajectory drift rather than scale ambiguity.
Scal3R consistently outperforms CUT3R in both settings, confirming that multi-reference relative pose querying improves not only relative pose accuracy but also global scale consistency.

\subsection{Comparison with Classic SLAM Systems}
\label{sec:supp-slam}

\begin{table*}[tb]
\centering
\caption{\textbf{Camera pose estimation results (ATE $\downarrow$) on KITTI with classic SLAM systems.}
Scal3R is competitive with calibrated SLAM and significantly outperforms calibration-free baselines without requiring camera intrinsics.}
\label{tab:kitti-ate}
\resizebox{\textwidth}{!}{%
\begin{tabular}{l c rrrrrrrrrrrr}
\toprule
\multirow{2}{*}{Method} & \multirow{2}{*}{Calib.} & \multicolumn{11}{c}{KITTI Sequence} & \multirow{2}{*}{Avg.} \\
\cmidrule(lr){3-13}
 & & 00 & 01 & 02 & 03 & 04 & 05 & 06 & 07 & 08 & 09 & 10 & \\
\midrule
ORB-SLAM2~\cite{mur2017orb} & \checkmark & 6.03 & 508.34 & 14.76 & 1.02 & 1.57 & 4.04 & 11.16 & 2.19 & 38.85 & 8.39 & 6.63 & 54.82 \\
DROID-SLAM~\cite{teed2021droid} & \checkmark & 170.60 & 91.03 & 255.22 & 1.25 & 0.35 & 59.79 & 32.07 & 14.03 & 138.76 & 55.83 & 13.45 & 75.67 \\
DROID-SLAM~\cite{teed2021droid} & \ding{55} & 190.93 & 89.50 & 239.93 & 9.27 & 0.37 & 133.13 & 131.22 & 70.33 & 144.53 & 187.18 & 165.44 & 123.80 \\
MASt3R-SLAM~\cite{murai2025mast3r} & \checkmark & 188.46 & 562.85 & 282.38 & 121.67 & 92.61 & 99.84 & 57.22 & 76.97 & 263.63 & 184.09 & 179.09 & 191.71 \\
MASt3R-SLAM~\cite{murai2025mast3r} & \ding{55} & 188.46 & 562.85 & 282.38 & 121.67 & 92.61 & -- & 57.22 & 76.97 & 263.63 & 184.09 & 179.09 & 200.90 \\
\midrule
Scal3R (CUT3R) & \ding{55} & 45.34 & 164.98 & 139.88 & 33.87 & 9.89 & 29.37 & 47.40 & 6.44 & 227.02 & 29.63 & 33.23 & 69.73 \\
Scal3R (STream3R) & \ding{55} & 57.90 & 176.43 & 170.75 & 10.86 & 9.30 & 39.05 & 18.07 & 15.19 & 173.83 & 73.91 & 33.87 & 70.83 \\
\bottomrule
\end{tabular}%
}
\end{table*}

\cref{tab:kitti-ate} compares Scal3R with classic SLAM systems on all eleven KITTI sequences.
ORB-SLAM2 achieves the lowest average ATE (54.82) when calibration is available, yet suffers catastrophic failure on seq.\ 01 (508.34), exposing its sensitivity to feature-poor highway scenes.
DROID-SLAM degrades substantially without calibration (75.67 $\to$ 123.80), and MASt3R-SLAM struggles across most sequences regardless of calibration, averaging over 190.
Without requiring any camera intrinsics, our CUT3R and STream3R variants achieve 69.73 and 70.83 respectively, competitive with calibrated DROID-SLAM and significantly outperforming all calibration-free baselines.
These results demonstrate that multi-reference relative pose querying with PGO can match or surpass established SLAM pipelines on long outdoor sequences while operating in a fully calibration-free online regime.

\subsection{Comparison with the Concurrent LongStream}
\label{sec:supp-longstream}

\begin{table*}[tb]
\centering
\caption{\textbf{Comparison with the concurrent LongStream~\cite{cheng2026longstreamlongsequencestreamingautoregressive} on KITTI (ATE $\downarrow$).}
LongStream attains lower absolute ATE by retraining a 1.3B backbone on large-scale data, while Scal3R reaches a comparable regime by adapting a frozen backbone with $\sim$1\% parameters and yields pairwise constraints that LongStream's single-reference design cannot provide.}
\label{tab:longstream}
\setlength{\tabcolsep}{6pt}
\resizebox{\textwidth}{!}{%
\begin{tabular}{lcccccc}
\toprule
Method & Pose formulation & Drift mitigation & KITTI ATE $\downarrow$ & Backbone & Train data / views & Compute \\
\midrule
LongStream~\cite{cheng2026longstreamlongsequencestreamingautoregressive} & single-reference & cache refresh & 51.9 & retrained, 1.3B & 14+ datasets, 10--80 views & $32\times$ A100, $>$3\,d \\
Scal3R (CUT3R) & multi-reference & PGO + loop closure & 69.7 & frozen, $\sim$1\% & TartanAir, 4 views & $1\times$ A100, 8\,h \\
\bottomrule
\end{tabular}%
}
\end{table*}

\cref{tab:longstream} situates Scal3R relative to LongStream~\cite{cheng2026longstreamlongsequencestreamingautoregressive}, a concurrent streaming method that also abandons first-frame global regression in favor of relative pose.
The two pursue orthogonal directions.
LongStream retrains a 1.3B VGGT-based backbone on large-scale driving and indoor data, predicting a single keyframe-relative pose per frame and suppressing drift through cache-consistent training with periodic cache refresh.
Scal3R instead keeps the backbone frozen and adds $\sim$1\% trainable parameters that query $K$ relative poses jointly in one forward pass.
This multi-reference formulation produces pairwise constraints that PGO and loop closure aggregate into a globally consistent trajectory, whereas LongStream's single-reference chain offers no such graph structure and cannot close loops.

The two methods occupy different points on the accuracy-versus-cost trade-off.
LongStream reaches a lower absolute ATE (51.9 vs.\ 69.7 on KITTI) by retraining a billion-scale backbone with 32 A100 GPUs for more than three days, while Scal3R attains a comparable regime by adapting a frozen backbone in 8 hours on a single GPU from only 4-view TartanAir samples.
They are thus complementary rather than competing: LongStream shows that a fully retrained backbone can push absolute accuracy, and Scal3R shows that the same long-sequence collapse can be resolved at the pose-interface level with a small fraction of the data and compute, while additionally enabling loop closure.

\subsection{Ablation: Asymmetric vs.\ Symmetric Attention Injection}
\label{sec:supp-attn-ablation}

\begin{table}[tb]
\centering
\caption{\textbf{Asymmetric vs.\ Symmetric attention injection (ATE $\downarrow$).}
Asymmetric injection is critical for outdoor sequences, reducing ATE by up to $10{\times}$.}
\label{tab:attn-ablation}
\setlength{\tabcolsep}{4pt}
\begin{tabular}{lrrrrr}
\toprule
\multirow{2}{*}{Method} & \multicolumn{3}{c}{Indoor} & \multicolumn{2}{c}{Outdoor} \\
\cmidrule(lr){2-4} \cmidrule(lr){5-6}
 & Sintel & TUM & ScanNet & vKITTI & KITTI \\
\midrule
Symmetric (VPT)   & \textbf{0.154} & 0.044 & \textbf{0.083} & 57.78 & 197.45 \\
Scal3R (Asymmetric)  & 0.168 & \textbf{0.033} & 0.092 & \textbf{5.63} & \textbf{69.73} \\
\bottomrule
\end{tabular}
\end{table}

\cref{tab:attn-ablation} compares symmetric attention injection via standard visual prompt tuning (VPT) with our asymmetric design.
On indoor benchmarks the two variants perform comparably, with symmetric injection slightly better on Sintel and ScanNet while asymmetric injection leads on TUM.
The critical difference emerges on outdoor sequences: asymmetric injection reduces ATE by an order of magnitude on vKITTI (57.78 $\to$ 5.63) and by nearly $3{\times}$ on KITTI (197.45 $\to$ 69.73).
Symmetric injection allows pose query tokens to attend to and be attended by all image tokens bidirectionally, which can dilute reference-specific geometric cues in long-range outdoor settings.
By restricting the information flow so that pose query tokens read from image features without modifying them (Eqs.~(2)--(3) in the main paper), asymmetric injection preserves the frozen backbone's representation quality and enables more accurate relative pose prediction at scale.

\subsection{Compatibility with Zero-Shot Methods}
\label{sec:supp-zeroshot}

\begin{table}[tb]
\centering
\caption{\textbf{Compatibility with zero-shot test-time training (ATE $\downarrow$).}
TTT3R consistently improves ATE when applied on top of Scal3R's globally optimized poses.}
\label{tab:zeroshot-compat}
\setlength{\tabcolsep}{4pt}
\begin{tabular}{lrrrrr}
\toprule
\multirow{2}{*}{Method} & \multicolumn{3}{c}{Indoor} & \multicolumn{2}{c}{Outdoor} \\
\cmidrule(lr){2-4} \cmidrule(lr){5-6}
 & Sintel & TUM & ScanNet & vKITTI & KITTI \\
\midrule
Scal3R          & 0.168 & 0.033 & 0.092 & 5.63 & 69.73 \\
Scal3R + TTT3R  & \textbf{0.154} & \textbf{0.028} & \textbf{0.064} & \textbf{4.66} & \textbf{62.05} \\
\bottomrule
\end{tabular}
\end{table}

\begin{figure}[tb]
    \centering
    \includegraphics[width=0.62\linewidth]{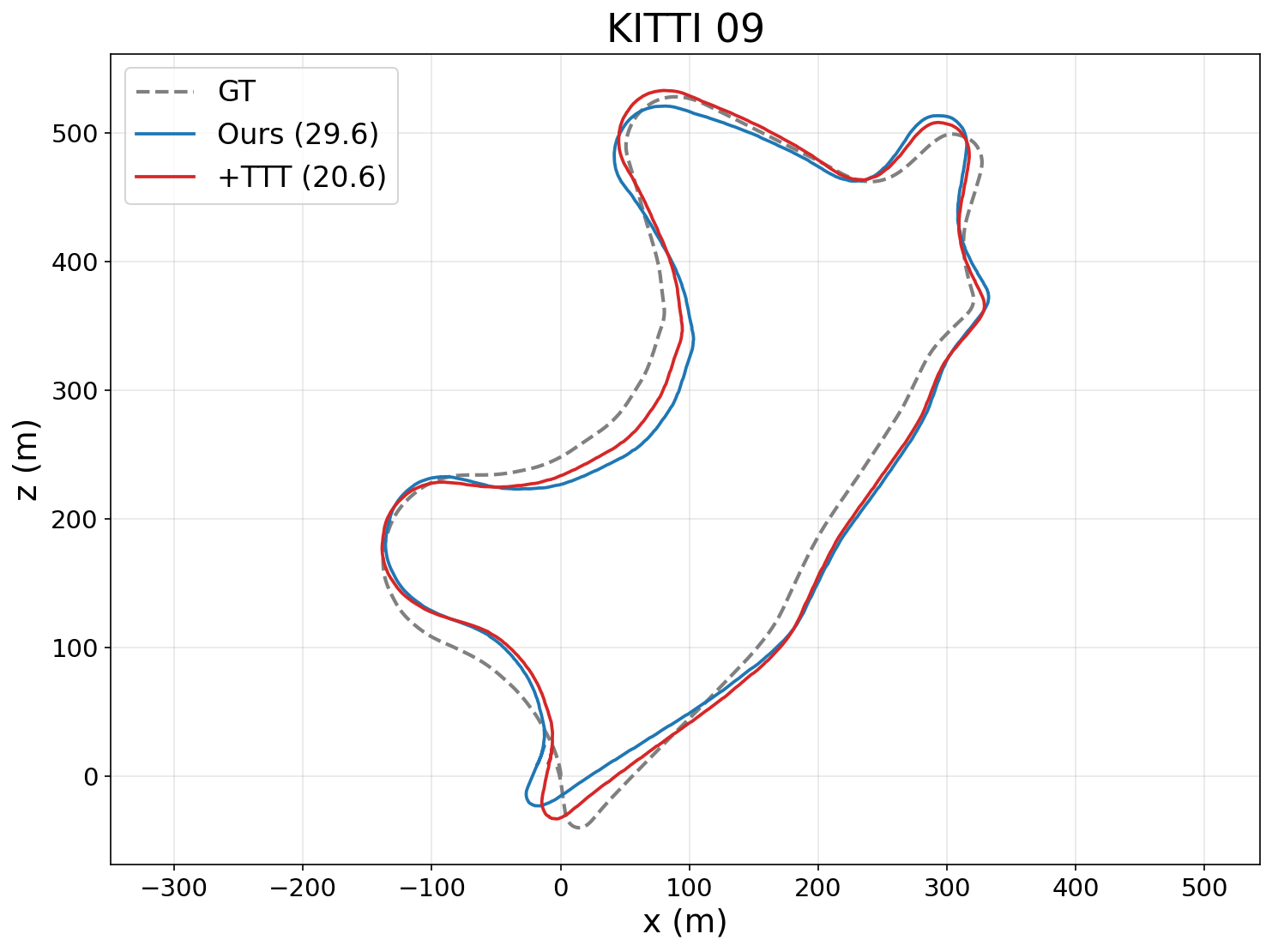}
    \caption{
    \textbf{Qualitative effect of test-time training on KITTI Seq.\ 09.}
    Adding TTT3R on top of Scal3R tightens the estimated trajectory against the ground truth, lowering ATE from 29.6 to 20.6.
    }
    \label{fig:ttt-compat}
\end{figure}

\cref{tab:zeroshot-compat} demonstrates that Scal3R is complementary to zero-shot test-time training methods such as TTT3R.
Applying TTT3R on top of our pipeline consistently improves ATE across all five benchmarks, with notable gains on ScanNet (0.092 $\to$ 0.064) and KITTI (69.73 $\to$ 62.05).
Since Scal3R produces globally optimized pose graphs, TTT3R benefits from higher-quality initial estimates compared to operating on raw sequential predictions.
This result confirms that our multi-reference relative pose querying and PGO pipeline serves as an effective foundation that can be further refined by orthogonal adaptation strategies at test time.
\cref{fig:ttt-compat} illustrates this complementarity on KITTI Seq.\ 09, where adding TTT3R tightens the estimated trajectory against the ground truth and lowers ATE from 29.6 to 20.6.

\subsection{Robustness to Dynamic Objects and Occlusions}
\label{sec:supp-dynamic}

\begin{figure*}[t]
    \centering
    \includegraphics[width=\textwidth]{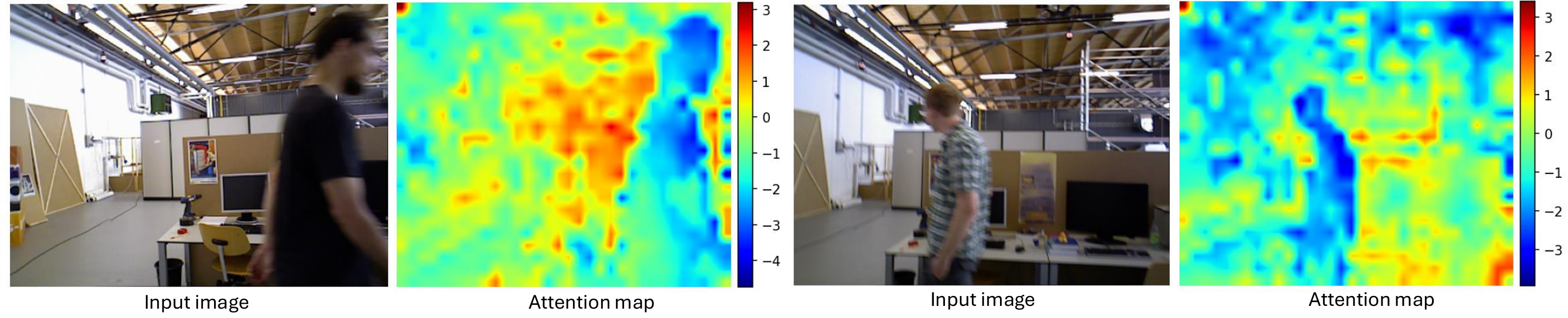}
    \vspace{-6mm}
    \caption{
    \textbf{Attention maps of relative pose query tokens on TUM-Dynamic.}
    The pose query tokens attend primarily to static structures while implicitly down-weighting dynamic pedestrians, without any explicit motion segmentation.
    }
    \label{fig:dynamic_atten_vis}
\end{figure*}

\cref{fig:dynamic_atten_vis} visualizes the attention maps of the relative pose query tokens $\{\tilde{\mathbf{q}}_k\}$ on TUM-Dynamic, where pedestrians occupy a significant portion of the frame.
The attention concentrates on static background structures such as desks, monitors, and walls, while assigning low activation to the moving persons in the foreground.
This behavior emerges naturally from the pose query tokens without any explicit dynamic-object mask or motion segmentation module, suggesting that the frozen backbone features already encode sufficient cues for the pose query tokens to distinguish static geometry from transient content.
The learned down-weighting of dynamic regions explains why Scal3R maintains accurate pose estimation on TUM-Dynamic despite large occlusions, and indicates that asymmetric attention injection provides an implicit robustness mechanism against scene dynamics.

\end{document}